\documentclass[lettersize,journal]{IEEEtran}

\usepackage{amssymb,amsmath}
\usepackage{cite}
\usepackage{graphicx}

\usepackage{psfrag}
\usepackage{url}
\usepackage[latin1]{inputenc}
\usepackage[absolute,overlay]{textpos}
\usepackage{gensymb}
\usepackage{cases}
\usepackage[font=footnotesize]{caption}
\usepackage{float}
\usepackage[linesnumbered,ruled,lined]{algorithm2e}
\newcommand{\argmax}[1]{{\operatorname{arg}\,\max_{#1}}\,}
\newcommand{\argmin}[1]{{\operatorname{arg}\,\min_{#1}}\,}
\usepackage{pgf}

\usepackage[nocomma]{optidef}

\usepackage{flushend}

\usepackage{dirtytalk}
\usepackage[printonlyused]{acronym}

\acrodef{MIMO}{multiple-input multiple-output}
\acrodef{mMIMO}{massive multiple-input multiple-output}
\acrodef{AWGN}{additive white Gaussian noise}
\acrodef{IA}{interference alignment}
\acrodef{SNR}{signal-to-noise ratio}
\acrodef{SINR}{signal-to-interference-plus-noise ratio}
\acrodef{CSI}{channel state information}
\acrodef{RL}{reinforcement learning}
\acrodef{5G}{fifth-generation}
\acrodef{6G}{sixth-generation}
\acrodef{ITU}{International Telecommunication Union}
\acrodef{MC}{massive communication}
\acrodef{HRLLC}{hyper-reliable low-latency communication}
\acrodef{VR}{virtual reality}
\acrodef{AR}{augmented reality}
\acrodef{AI}{artificial intelligence}
\acrodef{mMTC}{massive machine type communication}
\acrodef{URLLC}{ultra-reliable low-latency communication}
\acrodef{IA}{interference alignment}
\acrodef{TDMA}{time division multiple access}
\acrodef{FDMA}{frequency division multiple access}
\acrodef{CSI}{channel state information}
\acrodef{i.i.d.}{independent and identically distributed}
\acrodef{NLP}{natural language processing}
\acrodef{RNN}{recurrent neural network}
\acrodef{LSTM}{long short-term memory}
\acrodef{1D}{1-dimensional}
\acrodef{3D}{3-dimensional}
\acrodef{CNN}{convolutional neural network}
\acrodef{ReLU}{rectified linear unit}
\acrodef{tanh}{hyperbolic tangent}
\acrodef{IoE}{internet of everything}
\acrodef{FIR}{finite impulse response}
\acrodef{MSE}{mean squared error}
\acrodef{NMSE}{normalized mean squared error}
\acrodef{EBs}{Exabytes}
\acrodef{ML}{machine learning}
\acrodef{DL}{deep learning}
\acrodef{GRU}{gated recurrent unit}
\acrodef{Tx}{transmitter}
\acrodef{Rx}{receiver}
\acrodef{SVD}{singular value decomposition}
\acrodef{bps/Hz}{bits per second per Hertz}
\acrodef{UHD}{ultra high definition}
\acrodef{MDP}{Markov decision process}
\acrodef{DQN}{deep Q-networks}
\acrodef{DPG}{deterministic policy gradient}
\acrodef{DDPG}{deep deterministic policy gradient}
\acrodef{MADDPG}{multi-agent deep deterministic policy gradient}
\acrodef{DNN}{deep neural network}
\acrodef{MARL}{multi-agent reinforcement learning}
\acrodef{DRL}{deep RL}
\acrodef{V2V}{vehicle to vehicle}
\acrodef{HetNet}{heterogeneous network}
\acrodef{MDP}{Markov decision process}
\acrodef{UAV}{unmanned aerial vehicle}
\acrodef{mmWave}{millimeter-wave}
\acrodef{V2X}{vehicle to everything}
\acrodef{RSMA}{rate-splitting multiple access}
\acrodef{CTE}{centralized training and execution}
\acrodef{CTDE}{centralized training and decentralized execution}
\acrodef{DTE}{decentralized training and execution}
\acrodef{RL-Blind}{blind subspace coordination with RL}
\acrodef{RL-MaxSNR}{maximum SNR subspace coordination with RL}
\acrodef{Blind-ICISC}{blind inter-cell interference subspace coordination}
\acrodef{MaxSNR-ICISC}{maximum SNR inter-cell interference subspace coordination}
\acrodef{MMSE}{minimum mean square error}
\acrodef{TF-IA}{transformer-aided IA}
\acrodef{LSTM-IA}{LSTM-aided IA}
\acrodef{CNN-IA}{CNN-aided IA}
\acrodef{WMMSE}{weighted MMSE}
\acrodef{GFLOPs}{giga floating point operations}
\acrodef{FC}{fully connected}
\acrodef{DoF}{degrees-of-freedom}
\acrodef{THz}{terahertz}
\acrodef{IoT}{internet of things}
\acrodef{RF}{radio frequency}
\acrodef{OMP}{orthogonal matching pursuit}
\acrodef{ULA}{uniform linear array}
\acrodef{UPA}{uniform planar array}
\acrodef{UE}{user equipment}
\acrodef{AoD}{angle of departure}
\acrodef{AoA}{angle of arrival}
\acrodef{BS}{base station}
\acrodef{LS}{least squares}
\acrodef{LMMSE}{least minimum mean-squared error}
\acrodef{GenAI}{generative AI}
\acrodef{GAN}{generative adversarial network}
\acrodef{LoS}{line-of-sight}
\acrodef{NLoS}{non-line-of-sight}
\acrodef{FM}{Fundamentals of statistical signal processing: estimation
theory. }
\acrodef{ODE}{ordinary differential equation}
\acrodef{MAP}{maximum a posteriori}
\acrodef{PnP-PGD}{plug-and-play proximal gradient descent}
\acrodef{FACE}{flow-guided angular channel estimation}
\acrodef{QPSK}{quadrature phase shift keying}
\acrodef{SOMP}{simultaneous orthogonal matching pursuit}
\acrodef{CDF}{cumulative distribution function}
\acrodef{OT}{optimal transport}
\acrodef{CS}{compressed-sensing}
\acrodef{AMP}{approximate message passing}
\usepackage{subcaption}
\usepackage{stfloats}

\usepackage{amsthm}

\usepackage{booktabs}
\usepackage{color,soul}

\usepackage{textcomp}
\usepackage{bbm}
\usepackage{xcolor}
\def\BibTeX{{\rm B\kern-.05em{\sc i\kern-.025em b}\kern-.08em
    T\kern-.1667em\lower.7ex\hbox{E}\kern-.125emX}}

\usepackage{tabularx}
\usepackage{makecell}
\usepackage{multirow}

\begin{document}

\title{FACE: Flow-Guided Angular Channel Estimation Framework for mmWave MIMO Systems}
\author{
	\IEEEauthorblockN{Samitha Gunarathne, \textit{Graduate Student Member, IEEE}, Eslam Eldeeb, \textit{Member, IEEE}, Nurul Huda Mahmood, \textit{Member, IEEE} and Matti Latva-aho, \textit{Fellow, IEEE}
}

    \thanks{The authors are 
    with the Centre for Wireless Communications (CWC), University of Oulu, Finland. (e-mail: samitha.gunarathne@oulu.fi; eslam.eldeeb@oulu.fi; nurulhuda.mahmood@oulu.fi; matti.latva-aho@oulu.fi).}

    \thanks{This work was supported by the Research Council of Finland projects 6G Flagship (grant number: 369116), and ReWIN-6G (grant number: 357120), and by the Business Finland's 6GBridge project Local 6G (grant number: 8002/31/2022).}

}
\maketitle

\begin{abstract}
Acquiring accurate channel state information (CSI) is essential for millimeter-wave (mmWave) multi-input multi-output (MIMO) systems. However, it remains challenging due to high-dimensional channels, limited pilot observations, and measurement noise. To address these challenges, we propose flow-guided angular channel estimation (FACE), a novel channel estimation framework based on flow matching, an emerging generative modeling paradigm that learns continuous transport dynamics between probability distributions. FACE exploits the inherent angular-domain sparsity of mmWave channels and incorporates the learned flow matching model as a generative channel prior. Numerical results show that FACE consistently achieves higher channel estimation accuracy than both conventional estimators and diffusion-based generative estimator across diverse channel and system configurations. Moreover, FACE maintains high estimation accuracy with fewer pilot observations and achieves an average achievable rate close to the perfect-CSI benchmark. Compared with diffusion-based estimation, FACE requires substantially fewer sampling steps while achieving superior or comparable estimation accuracy, reducing inference time and computational cost. Overall, the proposed framework demonstrates the potential of flow matching for accurate, pilot-efficient, and computationally efficient mmWave MIMO channel estimation.

\end{abstract}


\section{Introduction}\label{sec: introduction}

\Ac{mmWave} and \ac{THz} communications have been recognized as key enablers for the \ac{6G} wireless communication systems~\cite{Diff_based_MIMO_BF, DiffPace}. By exploiting abundant bandwidth, these technologies offer promising solutions for supporting extremely high data rates~\cite{CS_techniques}. In particular, \ac{mmWave} communications can alleviate spectrum scarcity and capacity limitations while supporting emerging applications such as extended reality, the metaverse, massive \ac{IoT}, and ultra-high-definition video streaming~\cite{mmWave_THz_com}. Moreover, the short wavelengths of \ac{mmWave} signals allow a large number of antennas to be integrated into a compact physical space, thereby enabling large-scale antenna arrays and highly directional beamforming~\cite{mmWave_pros}.

Although a large number of antenna elements can be integrated into \ac{mmWave} \ac{MIMO} systems, hardware cost and power consumption constraints typically limit the number of available \ac{RF} chains~\cite{overview_of_mmWave_sigpro}. 
Typically, the number of \ac{RF} chains is significantly smaller than the number of antenna elements, creating a hybrid \ac{MIMO} architecture. This mismatch renders conventional channel estimation techniques designed for fully connected small-scale \ac{MIMO} impractical due to their high computational complexity and reliance on accurate channel statistics~\cite{overview_of_mmWave_sigpro, small_scale_MIMO_CE_1, small_scale_MIMO_CE_2, accurate_ce_mmWave, CE_using_CS}.  
This challenge is further exacerbated by the typically low pre-beamforming \ac{SNR} in \ac{mmWave} systems~\cite{CE_OMP}. Since the design of effective precoding matrices relies on accurate \ac{CSI}, reliable channel estimation is therefore essential to realize the potential performance gains of \ac{mmWave} \ac{MIMO} systems.

As the number of antenna elements increases, recovering high-dimensional \ac{CSI} from limited pilot measurements becomes an increasingly challenging ill-posed inverse problem~\cite{sig_pro_ill_posed_claim, DiffPace}. Conventional estimators, such as \ac{LS} and \ac{LMMSE} estimation, have been extensively used for channel estimation; however, their effectiveness becomes limited in large-scale \ac{MIMO} systems. In particular, \ac{LMMSE} relies on accurate channel statistics and incurs increased computational complexity as the channel dimension grows, limiting its practicality for large-scale \ac{mmWave} \ac{MIMO} systems~\cite{CE_gen_diff_models_for_ce}. 

To exploit the inherent sparsity of \ac{mmWave} channels in the angular-domain, \ac{CS}-based methods have been widely investigated for \ac{mmWave} channel estimation. In particular, greedy sparse recovery algorithms such as \ac{OMP} and \ac{SOMP} have been employed to estimate the dominant multi-path components of \ac{mmWave} \ac{MIMO} channels~\cite{CE_OMP}. However, these methods are susceptible to basis mismatch because the actual \acp{AoA} and \acp{AoD} may not align with the discretized angular grid used to construct the sparsifying dictionary, resulting in channel estimation performance degradation~\cite{basis_mismatch_CS}.

With recent advances in \ac{AI}, conventional \ac{DL}-based channel estimators typically formulate channel estimation as a supervised regression problem that directly maps the received pilot observations to the channel matrix~\cite{DL_estimator}. Although such discriminative models can provide accurate channel estimates, their performance may degrade when the channel conditions differ from those represented in the training data. 

Recently, \ac{GenAI} has emerged as a powerful paradigm for channel estimation due to its ability to learn the underlying distribution from complex high-dimensional data~\cite{CE_gen_diff_models_for_ce}, thereby providing a powerful prior for recovering channel distributions from limited and noisy observations~\cite{GenAI_survey}. Among these approaches, score-based and diffusion-based generative models~\cite{score_based_models, Denoising_diff_prob_models} have recently demonstrated promising performance for channel estimation. 
However, the iterative reverse-sampling process of diffusion-based methods typically requires a large number of sampling steps to obtain accurate channel estimates, resulting in considerable inference complexity. 

Recently, flow matching has emerged as an alternative generative method to diffusion models~\cite{FM_original}. Unlike diffusion models that learn stochastic paths between noise and data distributions, flow matching learns deterministic paths between the two distributions, which can simplify model training and sampling. Although flow matching has shown promise in other wireless domains such as multi-modal sensing~\cite{11623723}, semantic communication~\cite{emon2026flowsemflowmatchingadaptive}, and radio map reconstruction~\cite{jia2025radioflowefficientradiomap}, its potential for high-dimensional \ac{mmWave} \ac{MIMO} channel estimation remains largely unexplored.

Motivated by these limitations, we propose \ac{FACE}, a flow matching-based channel estimator for \ac{mmWave} systems. The proposed method relies on flow matching to learn a deterministic velocity field along probability paths between a simple noise distribution and the target channel distribution. This formulation enables accurate channel recovery with substantially fewer sampling steps than diffusion-based approaches. We integrate the trained flow matching model as a learned channel prior within a \ac{PnP-PGD} framework, combining measurement consistency with the learned channel distribution during inference. The main contributions of this work are summarized as follows:
\begin{itemize} 
    \item We formulate \ac{mmWave} \ac{MIMO} channel estimation as a \ac{MAP}-based regularized inverse problem that combines measurement-driven data-fidelity with a learned channel prior. By reformulating the estimation problem in the angular-domain, the proposed formulation exploits the inherent angular sparsity of \ac{mmWave} channels, providing a structured representation suitable for generative channel estimation.
    
    \item We propose a novel flow matching-based channel estimation framework, termed \ac{FACE}, which integrates a learned generative channel prior with \ac{PnP-PGD} for angular-domain \ac{mmWave} \ac{MIMO} channel estimation.
    
    \item We comprehensively evaluate the robustness of \ac{FACE} against the four baseline schemes across diverse system and channel configurations, including different antenna geometries, different antenna configurations, numbers of multi-path components, pilot lengths, and numbers of sampling steps.
    
    \item We demonstrate the computational efficiency of \ac{FACE}, achieving superior channel estimation accuracy with fewer sampling steps and lower computational cost than the diffusion-based estimator. Furthermore, \ac{FACE} achieves higher average achievable rates than the considered baseline schemes.
\end{itemize}

\subsection{Notations and Structure of the Paper} \label{subsec: notations}

Vectors and matrices are denoted by bold lowercase and uppercase letters, respectively. The $i$-th entry of a vector $\mathbf{x}$ is denoted by $\mathbf{x}_{i}$, while the element in the $i$-th row and $j$-th column of a matrix $\mathbf{X}$ is denoted by $\mathbf{X}_{i,j}$. The operator $(\cdot)^{\mathrm{H}}$ represents the conjugate transpose of a matrix. The $n$-dimensional identity matrix is denoted by $\mathbf{I}_n$. The sets of complex and real numbers are represented by $\mathbb{C}$ and $\mathbb{R}$, respectively. The real and imaginary parts of a complex number are denoted by $\textrm{Re}[\cdot]$ and $\textrm{Im}[\cdot]$, respectively. The notation $\mathcal{CN}(0,\sigma^{2})$ denotes a complex Gaussian distribution with zero mean and variance $\sigma^{2}$. The Frobenius norm and $l^{2}$ norm are denoted by $\|\cdot\|_{\textrm{F}}$, and $\|\cdot\|_{\textrm{2}}$ respectively. The expectation operator is represented by $\mathbb{E}\{\cdot\}$, and $\otimes$ denotes the Kronecker product.

 The remainder of this paper is organized as follows. Section~\ref{sec: prior_work} summarizes the related prior works. Section~\ref{sec: system_model} describes the considered \ac{mmWave} \ac{MIMO} system model, and the problem formulation. Section~\ref{sec: FACE_framework} presents the preliminaries of flow matching and proposed \ac{FACE} framework. Numerical results and an in-depth discussion are provided in Section~\ref{sec: numerical_results}. Lastly, Section~\ref{sec: Conclusion} concludes the paper.

\section{Prior Works} \label{sec: prior_work}

\ac{mmWave} channel estimation methods can be classified into three main categories, namely, conventional and sparsity-based approaches, \ac{DL}-based approaches and generative-based methods. This section reviews prior work on channel estimation in \ac{mmWave} \ac{MIMO} wireless communication systems.

\subsection{Conventional and Sparsity-based Estimation} \label{subsec: conventional CE}

The main idea of the \ac{CS}-based existing works is to exploit the inherent sparsity of the \ac{mmWave} channels in the angular-domain, where the channel paths reaching the receiver can be readily separated due to the highly directional propagation and scattering behavior of the \ac{mmWave} channel. 
In~\cite{CE_OMP}, the authors formulate a sparse signal recovery problem and develop a grid-based \ac{OMP} algorithm that employs a redundant dictionary of array response vectors for channel estimation in hybrid \ac{mmWave} \ac{MIMO} systems. The authors in~\cite{CE_joint_sparse_low_rank} propose a two-stage \ac{CS} technique for \ac{mmWave} channel estimation, leveraging joint sparse and low-rank structure of the channel. The proposed technique incurs lower sample complexity than conventional \ac{CS} approaches.

Reference~\cite{CE_using_LSE} proposes an iterative channel estimation method that employs \ac{LS} estimation and sparse message passing algorithm for \ac{mmWave} \ac{MIMO} systems. Exploiting the sparsity of the \ac{mmWave} channel, the iterative algorithm detects the non-zero values of the sparse channel matrix. The authors in~\cite{near_field_CE_for_XL_MIMO} propose a two-stage channel estimation strategy for a mixed \ac{LoS}/\ac{NLoS} near-field extremely-large \ac{MIMO} channel model. The \ac{LoS} path components are estimated iteratively by searching the coarse on-grid parameters, while the \ac{NLoS} path components are estimated by an \ac{OMP}-based algorithm. However, none of these studies evaluate their proposed methods using real channel measurements, leaving their performance under practical propagation conditions unverified.

\subsection{Deep Learning-based Channel Estimation} \label{subsec: learning-based CE}
 
\Ac{DL}-based channel estimation methods, such as ~\cite{CE_beamspace_ce_using_DL, CE_in_mMIMO_using_untrained_DNN, DL_based_CE}, have recently gained traction. The main idea is to leverage data-driven frameworks to learn the complex channel distribution from observed channel behaviour. The authors in~\cite{CE_beamspace_ce_using_DL} propose a complex-valued Gaussian mixture-learned \ac{AMP} framework for beamspace channel estimation, incorporating a learned shrinkage function into the \ac{AMP} algorithm to improve channel recovery. In~\cite{CE_in_mMIMO_using_untrained_DNN}, a \ac{DL}-based channel estimation scheme is developed for interference-limited multi-cell massive \ac{MIMO} systems. The proposed approach employs a specially designed \ac{DNN}-based deep image prior to denoise the received signal, followed by \ac{LS} channel estimation. The work in~\cite{DL_based_CE} develops a model-driven channel estimation approach by unfolding sparse Bayesian learning into a \ac{DNN}, where the network jointly optimizes the measurement matrix to improve estimation performance. These existing \ac{DL}-based channel estimation approaches do not evaluate their performance under varying numbers of multi-path components or validate their effectiveness using real channel measurements.

\subsection{Generative Model-based Estimation} \label{subsec: generative-based CE}

The work in~\cite{DiffPace} introduces a diffusion-based plug-and-play technique for channel estimation. The diffusion model captures the knowledge of the channel distribution and performs inference by solving an ordinary differential equation.
The work in~\cite{CE_gen_diff_models_for_ce} develops a novel diffusion model-based posterior inference technique for channel reconstruction. Furthermore, the proposed model recovers the \ac{CSI} from low-resolution quantized measurements. The authors in~\cite{CE_deep_gen_models} propose a \ac{CS} algorithm to acquire a preliminary channel estimate in extremely large-scale \ac{MIMO} systems by leveraging channel sparsity. Then, following the capabilities of \ac{GenAI}, the authors further propose a generative diffusion model-based channel refinement strategy. In~\cite{CE_Conditional_GANs}, the authors propose a conditional \ac{GAN} to model the channel effects. By incorporating the received pilot symbols as part of the conditioning information, the \ac{GAN} model generates specific samples for time-varying channels. However, these generative model-based channel estimation approaches lack comprehensive evaluations under varying numbers of multi-path components and different antenna array geometries. Furthermore, their estimation consistency across channel realizations has not been assessed through empirical \ac{CDF} comparisons with baseline methods.

The work in~\cite{Env_aware_CE_using_FM} proposes a flow matching-based \ac{CSI} acquisition framework for \ac{MIMO} systems that leverages multi-modal observations, including camera images, LiDAR point clouds, and GPS coordinates. Specifically, the authors employ cross-modal conditional flow matching to integrate environmental sensing information for accurate \ac{CSI} acquisition. However, the reliance on multiple sensing modalities may introduce additional sensing, processing, and communication overhead. In~\cite{FM_for_CE_and_Symbol_detection}, the authors develop a flow matching-based framework for joint channel estimation and symbol detection in massive \ac{MIMO} systems. The proposed training-free solver incorporates optimal-transport flow matching and employs a single pretrained score network to generate the channel flow, followed by Langevin-based refinement for payload detection. Existing studies on flow matching-based channel estimation have primarily focused on general \ac{MIMO} systems, with limited attention to the specific characteristics of \ac{mmWave} \ac{MIMO} channels. In particular, these approaches do not explicitly exploit the inherent angular-domain sparsity of \ac{mmWave} channels, which can provide a more structured representation for efficient channel estimation.

\section{System Model} \label{sec: system_model}

This section presents the details of the \ac{mmWave} \ac{MIMO} communication model, and the problem formulation.

\subsection{\Ac{mmWave} \ac{MIMO} System Model} \label{subsec: mmWave_system_model}
 
We consider a narrowband \ac{mmWave} \ac{MIMO} communication system equipped with $N_{\textrm{r}}$ receive and $N_{\textrm{t}}$ transmit antenna elements. 
For generality, we introduce the system model using the \ac{UPA}, where $N_{\textrm{t}}=N_{\textrm{t},x}N_{\textrm{t},z}$ and $N_{\textrm{r}}=N_{\textrm{r},x}N_{\textrm{r},z}$. Here, $N_{\textrm{t},x}$ and $N_{\textrm{t},z}$ denote the numbers of transmit antenna elements along the $x$- and $z$-axes, respectively, while $N_{\textrm{r},x}$ and $N_{\textrm{r},z}$ similarly denote the receive antenna elements. The \ac{ULA} geometry is obtained as a special case by setting $N_{\textrm{t},z}=N_{\textrm{r},z}=1$.
The transmitter employs $N_{\textrm{tRF}} \leq N_{\textrm{t}}$ \ac{RF} chains to transmit $N_{\textrm{s}}$ independent data streams simultaneously to the receiver. The receiver is equipped with $N_{\textrm{rRF}} \leq N_{\textrm{r}}$ \ac{RF} chains. We adopt a geometric channel model for the considered \ac{mmWave} \ac{MIMO} system, assuming $L$ propagation paths between the transmitter and receiver. The channel matrix is expressed as follows:
\begin{equation}
    \mathbf{H} = \sum_{l = 1}^{L} \alpha_{l} \mathbf{a}_{\textrm{r}}(\theta_{l}^{r}, \phi_{l}^{r}) \mathbf{a}_{\textrm{t}}^{\mathrm{H}}(\theta_{l}^{t}, \phi_{l}^{t}),  \label{eq: geometric_channel}
\end{equation}
where $L$ is the number of paths, $\alpha_{l} \sim \mathcal{CN}(0,1)$ is the complex path gain, $(\theta_{l}^{r}, \phi_{l}^{r})$ and $(\theta_{l}^{t}, \phi_{l}^{t})$ represent the \ac{AoA}, and \ac{AoD} pairs of receiver and transmitter for the $l$-th path respectively. The \ac{UPA} array response vectors at the receiver and transmitter are denoted by $\mathbf{a}_{\textrm{r}}(\theta_{l}^{r}, \phi_{l}^{r}) \in \mathbb{C}^{N_{\textrm{r}} \times 1}$ and $\mathbf{a}_{\textrm{t}}(\theta_{l}^{t}, \phi_{l}^{t}) \in \mathbb{C}^{N_{\textrm{t}} \times 1}$ respectively, and are expressed as:
\begin{align} \label{eq: rx_steeting_vector}
    \mathbf{a}_{\textrm{r}}(\theta_{l}^{r}, \phi_{l}^{r}) &=  \frac{1}{\sqrt{N_{\textrm{r}}}} \Big[e^{-j\frac{2\pi}{\lambda}d\sin{\theta_{l}^{r}} \cos{\phi_{l}^{r}} \mathbf{n}_{\textrm{r},x}} \otimes e^{-j\frac{2\pi}{\lambda}d\sin{\phi_{l}^{r}} \mathbf{n}_{\textrm{r},z}} \Big], \\
     \mathbf{a}_{\textrm{t}}(\theta_{l}^{t}, \phi_{l}^{t}) &=  \frac{1}{\sqrt{N_{\textrm{t}}}} \Big[e^{-j\frac{2\pi}{\lambda}d\sin{\theta_{l}^{t}} \cos{\phi_{l}^{t}} \mathbf{n}_{\textrm{t}, x}} \otimes e^{-j\frac{2\pi}{\lambda}d\sin{\phi_{l}^{t}} \mathbf{n}_{\textrm{t}, z}} \Big], \label{eq: tx_steeting_vector}
\end{align}
where $\lambda$ denotes the signal wavelength and $d=\lambda/2$ is the inter-element spacing. The antenna index vectors are defined as $\mathbf{n}_{\textrm{r},x} = [0, 1, \dots , N_{\textrm{r}, x}]$. $\mathbf{n}_{\textrm{r},z} = [0, 1, \dots , N_{\textrm{r}, z}]$, $\mathbf{n}_{\textrm{t},x} = [0, 1, \dots , N_{\textrm{t}, x}]$, and $\mathbf{n}_{\textrm{t}, z} = [0, 1, \dots , N_{\textrm{t}, z}]$. 

The transmitted signal can be expressed as:
\begin{equation}
    \mathbf{p} = \mathbf{F}_{\textrm{RF}}\mathbf{F}_{\textrm{BB}} \; \mathbf{s},  \label{eq: tx_signal}
\end{equation}
where $\mathbf{s} \in \mathbb{C}^{N_{\textrm{tRF}} \times 1}$ denotes the symbol vector, and $\mathbf{F}_{\textrm{RF}} \in \mathbb{C}^{N_{\textrm{t}} \times N_{\textrm{tRF}}}$ and $\mathbf{F}_{\textrm{BB}} \in \mathbb{C}^{N_{\textrm{s}} \times N_{\textrm{tRF}}}$ represent analog and digital precoders respectively. At the receiver terminal, the received signal can be expressed as follows:
\begin{equation}
    \mathbf{y} = \mathbf{H} \mathbf{p} + \mathbf{n},  \label{eq: rx_signal}
\end{equation}
where $\mathbf{n} \sim \mathcal{CN}(0,\,\sigma_{n}^{2} \mathbf{I}_{N_{\textrm{r}}})$ denotes the \ac{AWGN} vector with noise power $\sigma_{n}^{2}$. Let $\mathbf{P} = [\mathbf{p}_{1}, \mathbf{p}_{2}, \cdots , \mathbf{p}_{N_{\textrm{p}}}] \in \mathbb{C}^{N_{\textrm{t}} \times N_{\textrm{p}}}$ denotes the concatenated $N_{\textrm{p}}$ pilot matrix. The received observation signal matrix $\mathbf{Y} \in \mathbb{C}^{N_{\textrm{r}} \times N_{\textrm{p}}}$ can be expressed as follows:
\begin{equation}
    \mathbf{Y} = \mathbf{H} \mathbf{P} + \mathbf{N},  \label{eq:observation_matix}
\end{equation}
where $\mathbf{N} = [\mathbf{n}_{1}, \mathbf{n}_{2}, \cdots , \mathbf{n}_{N_{\textrm{p}}}] \in \mathbb{C}^{N_{\textrm{s}} \times N_{\textrm{p}}}$.

\subsection{Problem Formulation} \label{subsec: problem statement}

The objective is to recover $\mathbf{H}$ given the received signal $\mathbf{Y}$ in~\eqref{eq:observation_matix} and the pilot $\mathbf{P}$. However, due to the limited pilot observations and measurement noise, directly recovering $\mathbf{H}$ from $\mathbf{Y}$ in~\eqref{eq:observation_matix} constitutes an ill-posed inverse problem, necessitating the incorporation of appropriate prior information or structural knowledge to obtain an accurate channel estimate~\cite{sig_pro_ill_posed_claim}. 
We formulate the channel estimation problem as a \ac{MAP} problem as follows~\cite{MAP}:
\begin{equation}
    \hat{\mathbf{H}} = \argmax{\mathbf{H}} \log p(\mathbf{H} | \mathbf{Y}),
    \label{eq: MAP}
\end{equation}
where $\log p(\mathbf{H} | \mathbf{Y})$ denotes the log-posterior. According to Bayesian principles, the maximization in~\eqref{eq: MAP} can be reformulated as follows:
\begin{equation}
    \hat{\mathbf{H}} =  \argmax{\mathbf{H}} \log p(\mathbf{Y} | \mathbf{H}) + \log p(\mathbf{H}),
    \label{eq: MAP_reform}
\end{equation}
where $\log p(\mathbf{Y} | \mathbf{H})$ denotes the log-likelihood of the noisy received signal given the channel matrix, and $\log p(\mathbf{H})$ represents the log-prior distribution which consists the prior channel knowledge. 

The \ac{MAP} estimation in~\eqref{eq: MAP_reform} is a regularized maximization problem, where maximizing the log-likelihood term in~\eqref{eq: MAP_reform} is equivalent to minimizing the data-fidelity term~\cite{pattern_recognition}.
However, the log-prior term in~\eqref{eq: MAP_reform} is generally intractable in practice. Therefore, we adopt a regularizer term in place of $ \log p(\mathbf{H})$ that encapsulates the prior knowledge of the channel distribution~\cite{FM_CE_ICC_paper}. The optimization problem is formulated as follows:
\begin{equation}
    \mathbf{H}=\argmin{\mathbf{H}} \underbrace{{ \frac{1}{2\sigma_{n}^{2}} \| \mathbf{Y} - \mathbf{H} \mathbf{P} \|}_{\mathrm{F}}^{2}}_{\textrm{Data-fidelity term}} + \underbrace{ \mathcal{R}(\mathbf{H})}_{\substack{\text{Regularizer} \\ \text{term}}},
    \label{eq:regularized_min}
\end{equation}
where $\mathcal{R}(\mathbf{h})$ enforces the regularization term with prior knowledge of the ground truth channel distribution. 

In this work, we propose a generative framework for \ac{mmWave} channel estimation that employs a flow matching model to learn the underlying distribution of true channels as a channel prior, while incorporating the observed measurements during inference.

\section{FACE Framework for mmWave MIMO Channel Estimation} \label{sec: FACE_framework}

This section provides a detailed description of the proposed \ac{FACE} framework, which offers a systematic solution for acquiring \ac{CSI} using \ac{GenAI} approach. First, we present the preliminaries of flow matching, which are required to better present the proposed \ac{FACE} framework.

\subsection{Preliminaries on Flow Matching} \label{subsec: Preliminaries_of_FM}

Flow matching is a generative framework that learns a time-dependent vector field that transports samples from a simple prior distribution to a target data distribution~\cite{FM_original}. Let $p_{0}$ denote a tractable prior distribution (\emph{e.g.}, standard Gaussian distribution), and $p_{1}$ denote the target data distribution. The learned velocity field $v_{\theta}(\mathbf{x}_{t},t)$ defines a continuous probability flow between $p_0(\mathbf{x}_{t})$ and $p_1(\mathbf{x}_{t})$ over the normalized time interval $t\in[0,1]$. The resulting sample trajectories satisfy the following \ac{ODE}:
\begin{equation}
\frac{\mathrm{d}\mathbf{x}_{t}}{\mathrm{d}t} = v_{\theta}(\mathbf{x}_{t},t),
\end{equation}
where $\mathbf{x}_{0}\sim p_0(\mathbf{x}_{0})$. Unlike diffusion-based generative models, which rely on stochastic forward and reverse processes, flow matching directly learns the deterministic transport dynamics, thereby enabling efficient sample generation through numerical integration of the learned \ac{ODE}.

The key objective of flow matching is to learn the velocity field $v_{\theta}(\mathbf{x}_{t},t)$ that transports samples along the probability path from $p_{0}$ to $p_{1}$. During training, this learned velocity field is matched to a target velocity field $u_{t}(\mathbf{x}_{t})$ associated with the desired probability path. The neural velocity field $v_{\theta}(\mathbf{x}_{t},t)$ is trained to approximate $u_{t}(\mathbf{x}_{t})$ by minimizing the flow matching objective~\cite{FM_original}:
\begin{equation}
\mathcal{L}_{\textrm{FM}}(\theta)
=
\mathbb{E}_{t,,\mathbf{x}_{t}\sim p_{t}}
\left[ {\| v_{\theta}(\mathbf{x}_{t},t) - u_{t}(\mathbf{x}_{t}) \|}_{2}^{2} \right],
\label{eq: FM_original_loss}
\end{equation}
where $\theta$ denotes the learnable parameters of the neural velocity field. Minimizing~\eqref{eq: FM_original_loss} enables $v_{\theta}(\cdot)$ to approximate the target transport dynamics associated with the prescribed probability path. For Gaussian conditional probability paths, the \ac{OT} displacement interpolation induces straight-line transport trajectories between the paired source and target samples.

\subsection{Angular-Domain Conversion} \label{subsec: angular_domain}

In the \ac{FACE} framework, we preprocess the channels by converting them from the spatial-domain to the angular-domain. The angular-domain provides a more sparse representation of \ac{mmWave} channels, given that their multi-path propagation is dominated by a limited number of dominant paths $L$. The angular-domain conversion is represented as follows:
\begin{equation}
    \mathbf{H} = \mathbf{A}_{\textrm{r}} \mathbf{H}_{\textrm{a}} \mathbf{A}_{\textrm{t}}^{\mathrm{H}},  \label{eq: angular_domain_H}
\end{equation}
where $\mathbf{A}_{\textrm{r}} \in \mathbb{C}^{N_{\textrm{r}} \times N_{\textrm{r}}}$ and $\mathbf{A}_{\textrm{t}} \in \mathbb{C}^{N_{\textrm{t}} \times N_{\textrm{t}}}$ represent the discrete Fourier transform matrices for the array responses at the receiver and transmitter, respectively. The $\mathbf{H}_{\textrm{a}} \in \mathbb{C}^{N_{\textrm{r}} \times N_{\textrm{t}}}$ is the sparse angular-domain channel with its energy concentrated around a limited number of dominant angular components associated with the propagation paths.

\begin{figure}[t!]
    \centering

    \begin{subfigure}[t]{0.49\columnwidth}
        \centering
        \includegraphics[width=\linewidth, trim={0 0 0 0}, clip]
        {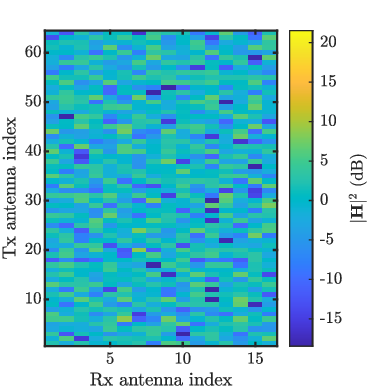}
        \caption{Spatial-domain channel $\mathbf{H}$.}
        \label{subfig:spatial_energy}
    \end{subfigure}
    \hfill
    \begin{subfigure}[t]{0.49\columnwidth}
        \centering
        \includegraphics[width=\linewidth, trim={0 0 0 0}, clip]
        {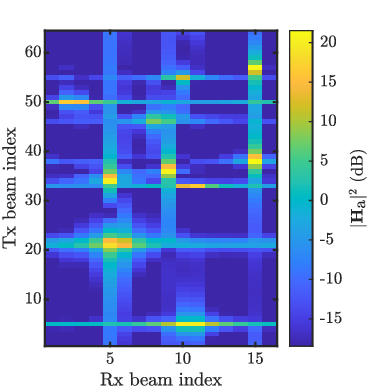}
        \caption{Angular-domain channel $\mathbf{H}_{\mathrm{a}}$.}
        \label{subfig:angular_energy}
    \end{subfigure}

    \caption{Normalized instantaneous channel power gain of the spatial- and angular-domain
    channels for $L=20$: (a) spatial-domain channel $\mathbf{H}$ and
    (b) angular-domain channel $\mathbf{H}_{\mathrm{a}}$.}
    \label{fig:energy_distribution}
\end{figure}

The angular-domain representation exploits the inherent sparsity of \ac{mmWave} propagation by concentrating channel energy into a small number of beam pairs corresponding to the dominant propagation paths. Compared with the dense spatial-domain representation, the angular-domain channel is sparser and more structured, thereby reducing the dimensionality of the matrix to be estimated, with most coefficients having negligible energy. This structure facilitates the learning process by allowing the flow matching model to primarily capture the evolution of a relatively small number of dominant channel components while maintaining the remaining coefficients close to zero. Consequently, although \ac{FACE} estimates the complete channel matrix, the number of significant coefficients that must be accurately modeled is substantially reduced, as discussed in the subsequent section. Fig.~\ref{fig:energy_distribution} illustrates the normalized instantaneous channel power gain of the spatial- and angular-domain channels for $L=20$ propagation paths. Compared with the spatial-domain representation, the angular-domain channel exhibits fewer dominant energy components, highlighting the inherent sparsity of \ac{mmWave} channels in the angular domain.

\subsection{FACE Training} \label{subsec: FM_CE_method_training}

\ac{FACE} is a generative framework for \ac{mmWave} channel estimation, which incorporates a flow matching model to learn a prior over the distribution of the true channels $\mathbf{H}_{\textrm{a}}$, while conditioning on the measurement during inference. This allows a single trained prior to be reused across different pilot realizations and \ac{SNR} conditions without retraining. As discussed in Section~\ref{subsec: angular_domain}, we incorporate the inherent sparsity of the angular-domain representation to reformulate the received signal as follows:
\begin{equation}
    \mathbf{Y}_{\textrm{a}} = \mathbf{H}_{\textrm{a}} \mathbf{P} + \mathbf{N},  \label{eq: angularsomain_rx_signal}
\end{equation}
where $\mathbf{Y}_{\textrm{a}}$ denotes the angular-domain received signal. Due to the sparsity of $\mathbf{H}_{\textrm{a}}$, the angular-domain received signal $\mathbf{Y}_{\textrm{a}}$ is sparser compared to the spatial-domain received signal $\mathbf{Y}$. Therefore, we reformulate the \ac{MAP} estimation problem~\eqref{eq:regularized_min} as follows:
\begin{equation}
    \hat{\mathbf{H}}_{\textrm{a}} =\argmin{\mathbf{H}_{\textrm{a}}}{ \frac{1}{2\sigma_{n}^{2}} \| \mathbf{Y}_{\textrm{a}} - \mathbf{H}_{\textrm{a}} \mathbf{P} \|}_{\mathrm{F}}^{2} + \mathcal{R}(\mathbf{H}_{\textrm{a}}), \label{eq: reform_regularized_LS}
\end{equation}
where $\mathcal{R}(\mathbf{H}_{\textrm{a}})$ is a regularization term with prior knowledge of the ground truth angular-domain channel distribution.

\begin{figure*}[t!]
    \centering
    \includegraphics[width=2\columnwidth,trim={0 0 0 0},clip]{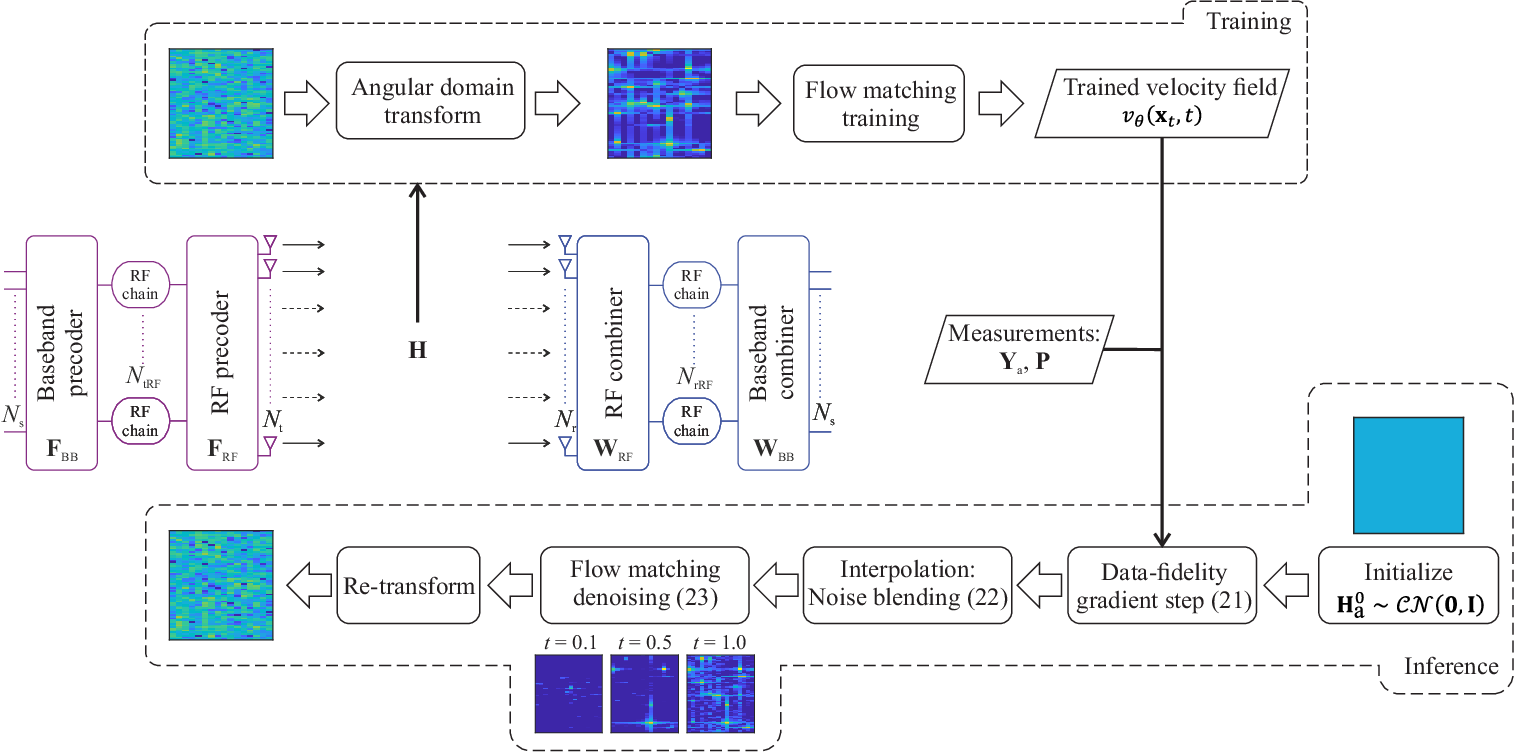}
    \caption{Proposed \Ac{FACE} framework for \ac{mmWave} channel estimation.}
    \label{fig:FACE_diagram}
\end{figure*}

Let the set of training \ac{mmWave} spatial-domain channel matrices be $\mathcal{H} = \{\mathbf{H}_{1}, \mathbf{H}_{2}, \cdots , \mathbf{H}_{M}\}$, where $M$ is the training sample size. Each training channel matrix is transformed into an angular-domain representation using~\eqref{eq: angular_domain_H}. The set of angular-domain training channel matrices is $\mathcal{H}_{\textrm{a}} = \{\mathbf{H}_{{\textrm{a}}, 1}, \mathbf{H}_{{\textrm{a}}, 2}, \cdots , \mathbf{H}_{{\textrm{a}}, M}\}$. We represent the corresponding angular-domain channels as real-valued tensors by applying the stacking operator $\mathcal{S} : \mathbb{C}^{N_{\textrm{r}} \times N_{\textrm{t}}} \mapsto \mathbb{R}^{2 \times N_{\textrm{r}} \times N_{\textrm{t}}}$ on the angular-domain channels as follows:
\begin{equation}
    \tilde{\mathbf{H}}_{\textrm{a}, i} = \mathcal{S}(\mathbf{H}_{\textrm{a}, i}) \triangleq \big[ \mathrm{Re}[\mathbf{H}_{\textrm{a}, i}], \mathrm{Im}[\mathbf{H}_{\textrm{a}, i}] \big]^{\mathrm{T}}. \label{eq: stacking}
\end{equation}

\Ac{FACE} learns a time-dependent velocity field $v_{\theta} (\cdot)$ that maps a probability path from the noise distribution to the target angular-domain channel distribution. We consider the simple linear path interpolation between the two distributions as follows:
\begin{equation}
    \mathbf{x}_{t} = (1 - t)\mathbf{x}_{0} + t\mathbf{x}_{1},
    \label{eq: interpolation}
\end{equation}
where $\mathbf{x}_{0} \sim \mathcal{N}(\mathbf{0}, \mathbf{I})$, $\mathbf{x}_{1} \sim \{ \tilde{\mathbf{H}}_{\textrm{a}, i} \}_{i = 1}^{M}$, and $t \sim \mathrm{Unif}[0, 1]$. During training, the proposed \ac{FACE} optimizes the velocity field parameters by training the following loss:
\begin{equation}
    \mathcal{L}(\theta) = \mathbb{E}_{\mathbf{x}_{0},\mathbf{x}_{1},t}\Big[ \Big\| v_{\theta} \big( \mathbf{x}_{t}, t \big) - \big( \mathbf{x}_{1}- \mathbf{x}_{0} \big) \Big\|_2^2 \Big],
    \label{eq: FM_loss}
\end{equation}
where $\mathbf{x}_{1} - \mathbf{x}_{0}$ is the target velocity field.
The proposed model learns a marginal distribution over the angular-domain training channels in an unsupervised manner and independent of the pilot pattern. Therefore, the resulting prior generalizes across pilot densities and \ac{SNR} conditions without re-training. Since $\mathbf{H}_{\textrm{a}}$ concentrates most of the channel's energy in a small number of entries, the effective support of the training distribution is lower-dimensional than $\mathbf{H}$ as empirically evaluated in Section~\ref{sec: numerical_results}. Algorithm~\ref{algo:FM_CE_training} presents the training of the proposed \ac{FACE} framework.

\subsection{FACE Sampling} \label{subsec: FM_CE_method_inference}

To incorporate the learned flow matching model as a channel prior during inference, \ac{FACE} constructs a time-dependent denoiser from the learned velocity field~\cite{Proximal_denoiser}. The optimal flow-matching velocity is given by:
\begin{equation}
\mathbf{v}^{\star}(\mathbf{x}_t,t) = \mathbb{E}\!\left[ \mathbf{x}_1-\mathbf{x}_0 \mid \mathbf{x}_t \right].
\end{equation}
Recall the linear path interpolation presented in~\eqref{eq: interpolation} which can be rewritten as:
\begin{equation}
\mathbf{x}_t + (1-t)(\mathbf{x}_1-\mathbf{x}_0) = \mathbf{x}_1.
\end{equation}
Taking the conditional expectation, we obtain:
\begin{equation}
\mathbf{x}_t + (1-t)\mathbf{v}^{\star}(\mathbf{x}_t,t) = \mathbb{E}\!\left[ \mathbf{x}_1\mid\mathbf{x}_t
\right].
\end{equation}
Accordingly, we define the flow-based denoiser as follows:
\begin{equation}
D_{t}(\mathbf{x}_{t})
=
\mathbf{x}_{t}+(1-t)v_{\theta}(\mathbf{x}_{t},t)
\approx \mathbf{x}_{1}.
\label{eq: denoiser2}
\end{equation}
Therefore, $D_t(\mathbf{x}_{t})$ can be interpreted as an estimate of the clean endpoint associated with the current intermediate state.

\begin{algorithm}[!t]

\textbf{Input:} Training dataset $\mathcal{H}$, number of epochs $N_{\textrm{e}}$, and number of steps per epoch $N_{\textrm{steps}}$

Convert training dataset $\mathcal{H}$ to $\mathcal{H}_{\textrm{a}}$ using~\eqref{eq: angular_domain_H} \\
Initialize neural network $v_{\theta}$ and Gaussian source distribution $\mathbf{x}_{0} \sim\mathcal{N}(\mathbf{0}, \mathbf{I})$

\For{episode $e \in \{1, \dots, N_{e}\}$}{

\For{step $s \in \{1,\dots,N_{\textrm{steps}}\}$}{

Sample a batch $\mathbf{x}_{1}$ from the training dataset $\mathcal{H}_{\textrm{a}}$ and sample $\mathbf{x}_{0} \sim\mathcal{N}(\mathbf{0}, \mathbf{I})$

Sample $t \sim \mathrm{Unif}[0, 1]$

Compute $\mathbf{x}_{t} = (1 - t)\mathbf{x}_{0} + t\mathbf{x}_{1}$

Compute the training loss using~\eqref{eq: FM_loss}

Update $\theta$ by applying gradient descent on the training loss
}
}
\textbf{Output:} Trained velocity field $v_{\theta}(\mathbf{x}_{t}, t)$
\caption{Training of the \ac{FACE} framework for \ac{mmWave} \ac{MIMO} channel estimation.}
\label{algo:FM_CE_training} \vspace{0mm}
\end{algorithm}


To recover the channel during inference, the proposed \ac{FACE} framework employs an iterative \ac{PnP-PGD} scheme to solve the regularized inverse problem in~\eqref{eq: reform_regularized_LS}~\cite{Proximal_denoiser, PnP_methods}. The \ac{PnP-PGD} procedure combines pilot-measurement information with the channel prior learned by the flow matching model. Specifically, each iteration consists of a data-fidelity gradient step that updates the current channel estimate to improve its consistency with the observed pilot measurements, followed by a flow matching-based denoising step that progressively guides the updated estimate toward the learned channel distribution. The inference procedure is performed over $K$ uniformly spaced time steps in $t\in[0,1]$, with $t^{(k)}=k/K$. At each time step $t^{(k)}$, the received pilot matrix $\mathbf{Y}_{\textrm{a}}$ and transmitted pilot matrix $\mathbf{P}$ are incorporated through the data-fidelity term in~\eqref{eq: reform_regularized_LS}. Accordingly, the gradient descent update is expressed as:
\begin{equation}
    \mathbf{Z}^{(k + 1)} = \mathbf{H}_{\textrm{a}}^{(k)} + \frac{\gamma_{k}}{\sigma_{n}^{2}} \big(\mathbf{Y}_{\textrm{a}} - \mathbf{H}_{\textrm{a}}^{(k)} \mathbf{P} \big) \mathbf{P}^{\mathrm{H}},
    \label{eq: gradient_decent}
\end{equation}
where $\gamma_{k}$ denotes the step size. After the data-fidelity step, the flow matching model guided denoising step is applied as:
\begin{align}
    \tilde{\mathbf{Z}}^{(k + 1)} &= \mathcal{S} \big( (1 - t^{(k)}) \mathbf{\varepsilon}^{(k)} + t^{(k)} \mathbf{Z}^{(k + 1)}\big), \label{eq: denoising_step} \\
    \mathbf{H}_{\textrm{a}}^{(k + 1)} &= \mathcal{S}^{-1} \big( D_{{t}^{(k)}} (\tilde{\mathbf{Z}}^{(k + 1)})\big),
    \label{eq: H_update}
\end{align}
where $\mathbf{\varepsilon}^{(k)} \in \mathbb{C}^{N_{\textrm{r}} \times N_{\textrm{t}}}$ denotes the \ac{i.i.d.} entries following $\mathcal{CN}(\mathbf{0}, \mathbf{I})$ and $\mathcal{S}^{-1}$ denotes the inverse transformation of~\eqref{eq: stacking}. The evolution of channel reconstruction during inference is further discussed in Section~\ref{sec:genai_analysis}. Algorithm~\ref{algo: FM_CE_inference} summarizes the complete procedure of the flow matching model inference for channel estimation. Fig.~\ref{fig:FACE_diagram} illustrates the overall pipeline of the proposed \ac{FACE} framework.

\begin{algorithm}[!t]

\textbf{Input:} trained velocity field $v_{\theta}$, pilot $\mathbf{P}$, received signal $\mathbf{Y}_{\textrm{a}}$, and number of iterations $K$

Initialize $\mathbf{H}_{\textrm{a}}^{0} \sim \mathcal{N}(\mathbf{0}, \mathbf{I})$;

\For{step $k \in \{1, \dots, K\}$}{

Perform gradient decent step in data-fidelity term using~\eqref{eq: gradient_decent}

Update $\mathbf{H}_{\textrm{a}}^{(k + 1)}$ using~\eqref{eq: denoising_step} and~\eqref{eq: H_update}
}

Perform inverse transformation on estimated $\hat{\mathbf{H}}_{\textrm{a}}$ to get spatial-domain channel $\hat{\mathbf{H}}$ using~\eqref{eq: angular_domain_H}

\textbf{Output:} Estimated channel matrix $\hat{\mathbf{H}}$

\caption{Inference of the \ac{FACE} framework for \ac{mmWave} \ac{MIMO} channel estimation.}
\label{algo: FM_CE_inference} \vspace{0mm}
\end{algorithm}
 \vspace{-1mm}

\section{Numerical results} \label{sec: numerical_results}

This section presents the numerical evaluations of the proposed \ac{FACE} framework for \ac{mmWave} \ac{MIMO} channels. First, we provide a detailed description of the simulation setup and the baseline schemes considered. Afterwards, we evaluate the proposed \ac{FACE} in terms of key metrics, including channel estimation accuracy across a wide variety of scenarios, average achievable rate, and computational complexity.

\subsection{Simulation Setup and Dataset} \label{sec: simulation_setup}

We generate a \ac{mmWave} channel dataset by employing the geometric channel model in~\eqref{eq: geometric_channel} at a carrier frequency of $f_{\textrm{c}} = 60$ GHz. We construct the symbol vectors by randomly selecting \ac{QPSK} symbols. The cascaded analog-digital precoder $\mathbf{F}_{\mathrm{RF}}\mathbf{F}_{\mathrm{BB}}$ is constructed using random phase shifts, with the analog precoder $\mathbf{F}_{\mathrm{RF}}$ satisfying the constant-modulus constraint. The \acp{AoA} $\{\theta_{l}^{r}, \phi_{l}^{r}\}_{l=1}^{L}$ and \acp{AoD} $\{\theta_{l}^{t}, \phi_{l}^{t}\}_{l=1}^{L}$ are independently and uniformly sampled from the interval $[-\frac{\pi}{2}, \frac{\pi}{2}]$. We consider train and test datasets consisting of $10,000$ and $1,000$ channel realizations, respectively.

In addition, to evaluate the proposed \ac{FACE} framework under realistic propagation conditions, we use a $60$ GHz channel measurement dataset~\cite{mmWave_dataset}. The dataset contains channel-sounding measurements collected at multiple receiver positions across three measurement sites. For each receiver position, the multi-path components are characterized by measured parameters, including the path delay, \ac{AoA}, \ac{AoD}, and received power. These measurement-derived multi-path parameters are used to construct the corresponding \ac{MIMO} channel matrices for a \ac{ULA} configuration with $N_{\textrm{t}}=32$ and $N_{\textrm{r}}=16$. The mean and median numbers of multi-path components in the dataset are $100.4$ and $99$, respectively. From the resulting channel realizations, we use $8,000$ samples for training and reserve $1,000$ for testing.

\begin{table}[!h]
\centering
 \caption{Simulation parameters}
\label{table: simulation-params}
\begin{tabular}{@{}ll|ll@{}}
\toprule
\multicolumn{2}{c|}{\textbf{\ac{mmWave} channel}} & \multicolumn{2}{c}{\textbf{\ac{FACE} framework }} \\
\midrule
\textbf{Parameter} & \textbf{Value} & \textbf{Parameter} & \textbf{Value} \\ 
\midrule
$f_{\textrm{c}}$ & $60$ GHz & Architecture & U-Net \\
$N_{\textrm{t}}$ & $\{32, 64, 128\}$ & Optimizer & Adam \\
$N_{\textrm{r}}$ & $\{16, 32\}$ & Residual blocks & 2 \\
$N_{\textrm{tRF}}$ & $16$ & Batch size & 32 \\
$L$ & $\{3, 10, 50\}$ & Learning rate & $1 \times 10^{-4}$ \\
$N_{\textrm{p}}$ & $\{20, 50, 64, 100\}$ &  Up/down sampling & convolutional \\
SNR (dB) & $\{-10, -5, 0, 5, $ & Epochs & $400$ \\
 & $10, 15, 20, 25\}$ & $\gamma_{k}$ & $0.095$ \\
\bottomrule
\end{tabular}
\end{table}

Table~\ref{table: simulation-params} summarizes the channel generation parameters, network configuration, and training hyperparameters used in the subsequent performance evaluations. The proposed \ac{FACE} framework employs a flow matching model parameterized by a U-Net-based architecture~\cite{Unet}, consisting of a downsampling path (\emph{i.e.,} encoder) and a symmetric upsampling path (\emph{i.e.,} decoder) connected through skip connections, with residual blocks incorporated at each resolution level. As described in Section~\ref{sec: FACE_framework}, the network takes the real and imaginary components of the channel matrix as input, with spatial dimensions of $N_{\textrm{t}} \times N_{\textrm{r}}$ antenna configuration. The model is trained for $400$ epochs on the generated training dataset using PyTorch framework on a single NVIDIA GH200 GPU.

\subsection{Baseline Schemes} \label{sec: baselines}

We consider four main baseline schemes as follows:

\begin{itemize}
    \item \textbf{Diffusion}: Diffusion models are a class of \ac{GenAI} models that generate samples from a Gaussian prior through an iterative denoising process. We follow the formulation in~\cite{CE_gen_diff_models_for_ce} to implement the diffusion-based channel estimator. To ensure a fair comparison, the diffusion-based channel estimation baseline employs the same U-Net backbone architecture as the proposed \ac{FACE} framework;
    \item  \textbf{\Ac{LMMSE}}: The \ac{LMMSE} estimator provides the optimal linear estimate by minimizing the Bayesian \ac{MSE} between the estimated and true channel matrices~\cite{CE_LMMSE};
    \item \textbf{Orthogonal matching pursuit}: \Ac{OMP} is a classical greedy \ac{CS} algorithm for channel estimation that exploits the inherent sparsity of \ac{mmWave} channels in the angular-domain~\cite{CE_OMP}.
    \item  \textbf{Simultaneous orthogonal matching pursuit}: \Ac{SOMP} extends \ac{OMP}'s greedy sparse-recovery principle to jointly recover the channel from multiple measurement vectors by exploiting both angular-domain sparsity and a common sparse support across measurements~\cite{accurate_ce_mmWave}.
\end{itemize}

\subsection{Evaluation Metrics} \label{sec: metrics}

We consider the following evaluation metrics in the simulations:
\begin{itemize}
    \item \textbf{\Ac{NMSE}:} We employ the \ac{NMSE} as the primary performance metric, expressed as:
    \begin{equation}
    \label{eq: NMSE}
        \textrm{NMSE}(\mathbf{H}, \hat{\mathbf{H}}) = 10\log_{10}\bigg( \mathbb{E}\bigg[ \frac{\| \mathbf{H} - \hat{\mathbf{H}} \|^{2}_{\textrm{F}}}{\| \mathbf{H} \|^{2}_{\textrm{F}}} \bigg] \bigg),
    \end{equation}
    where $\mathbf{H}$ and $\hat{\mathbf{H}}$ denote the true and estimated channel matrices, respectively. The \ac{NMSE} quantifies the channel estimation accuracy and is used to compare the performance of \ac{FACE} with the considered baseline schemes. 

    \item \textbf{Average achievable rate:} To assess the impact of channel estimation accuracy on communication performance, we evaluate the average achievable rate when the transmit precoder is designed from the estimated channel $\hat{\mathbf{H}}$. Let the \ac{SVD} of the estimated channel be $\hat{\mathbf{H}}=\hat{\mathbf{U}}\hat{\mathbf{\Sigma}}\hat{\mathbf{V}}^{\mathrm{H}}$, where $\hat{\mathbf{U}}$ and $\hat{\mathbf{V}}$ contain the left and right singular vectors, respectively, and $\hat{\mathbf{\Sigma}}$ contains the corresponding singular values. The transmit precoder is constructed using the $N_{\textrm{s}}$ dominant right singular vectors as $\mathbf{F}=\hat{\mathbf{V}}(:,1:N_{\textrm{s}})$. The resulting effective channel is $\mathbf{H}_{\textrm{eff}}=\mathbf{H}\mathbf{F}$, where $\mathbf{H}$ denotes the true channel. The achievable rate is expressed as follows:
    \begin{equation}
        R = \log_{2}\det \bigg( \mathbf{I}_{N_{\textrm{r}}}+ \frac{\mathrm{SNR}}{N_{\textrm{s}}} \mathbf{H}_{\textrm{eff}} \mathbf{H}_{\textrm{eff}}^{\mathrm{H}}\bigg),
    \end{equation}
    The average achievable rate is obtained by averaging $R$ over all channel realizations in the test set.

    \item \textbf{Computational complexity:} We evaluate the computational efficiency of \ac{FACE} and the diffusion-based estimator through theoretical complexity analysis and \ac{GFLOPs}-based computation. Specifically, we report the \ac{GFLOPs} per sampling step and assess the overall inference cost based on the number of sampling steps required by each method.
    
\end{itemize}
In addition, we examine the scalability of the proposed method and the baselines while sweeping key parameters. We present the \ac{NMSE} performance as we vary the number of antennas, the number of multi-paths, the pilot length, the sampling step size, and the inference time.

\subsection{NMSE Results} \label{sec:nmse_results}


\begin{figure*}[t!]
    \centering

    \begin{subfigure}[t]{0.32\textwidth}
        \centering
        \includegraphics[width=1.1\linewidth,trim={5pt 2pt 10pt 10pt},clip]{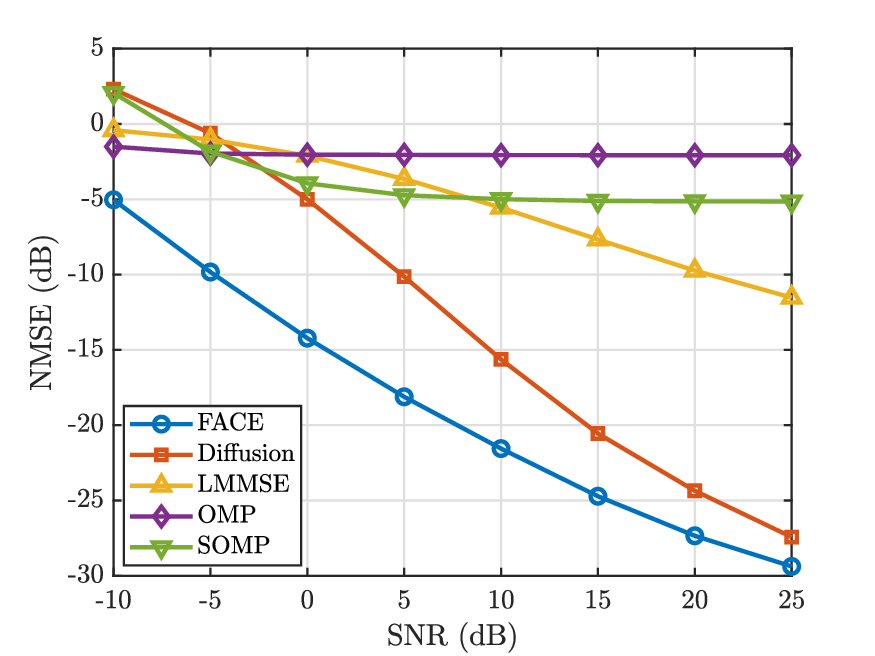}
        \caption{$N_{\textrm{t}} = 32$, $N_{\textrm{r}} = 16$}
        \label{fig:ULA_antennas_32x16}
    \end{subfigure}
    \hfill
    \begin{subfigure}[t]{0.32\textwidth}
        \centering
        \includegraphics[width=1.1\linewidth,trim={5pt 2pt 10pt 10pt},clip]{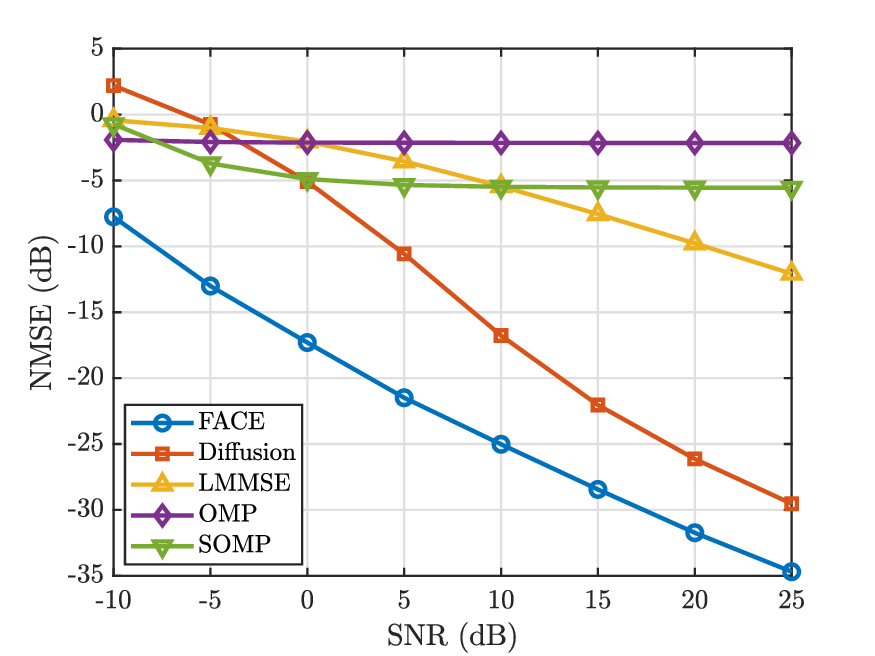}
        \caption{$N_{\textrm{t}} = 64$, $N_{\textrm{r}} = 16$}
        \label{fig:ULA_antennas_64x16}
    \end{subfigure}
    \hfill
    \begin{subfigure}[t]{0.32\textwidth}
        \centering
        \includegraphics[width=1.1\linewidth,trim={5pt 2pt 10pt 10pt},clip]{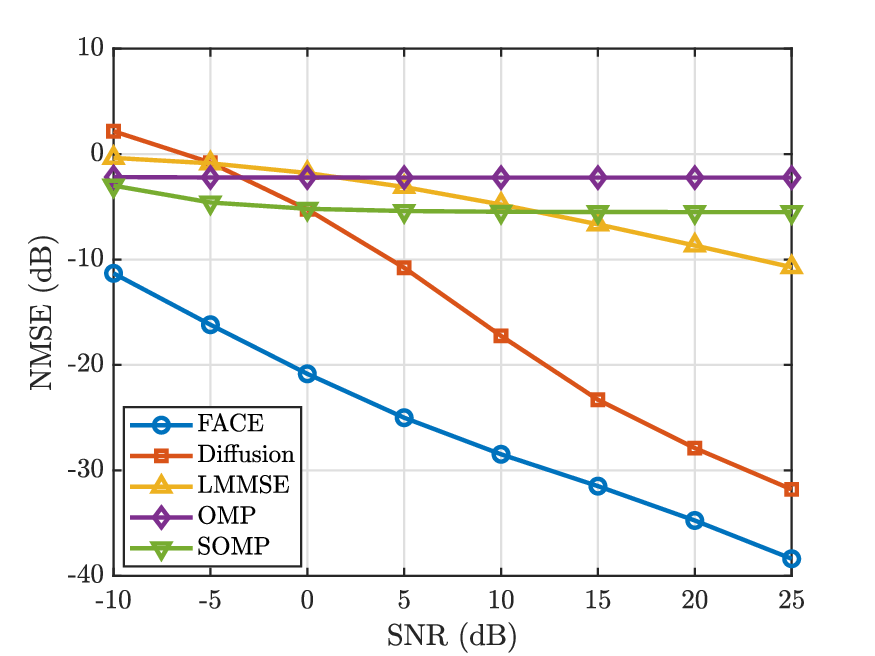}
        \caption{$N_{\textrm{t}} = 128$, $N_{\textrm{r}} = 32$}
        \label{fig:ULA_antennas_128x32}
    \end{subfigure}

    \caption{Variation of \ac{NMSE} with \ac{SNR} for different \ac{ULA} geometry.}
    \label{fig:ULA_antenna_configurations}
\end{figure*}

\begin{figure*}[t!]
    \centering

    \begin{subfigure}[t]{0.32\textwidth}
        \centering
        \includegraphics[width=1.1\linewidth,trim={5pt 2pt 10pt 10pt},clip]{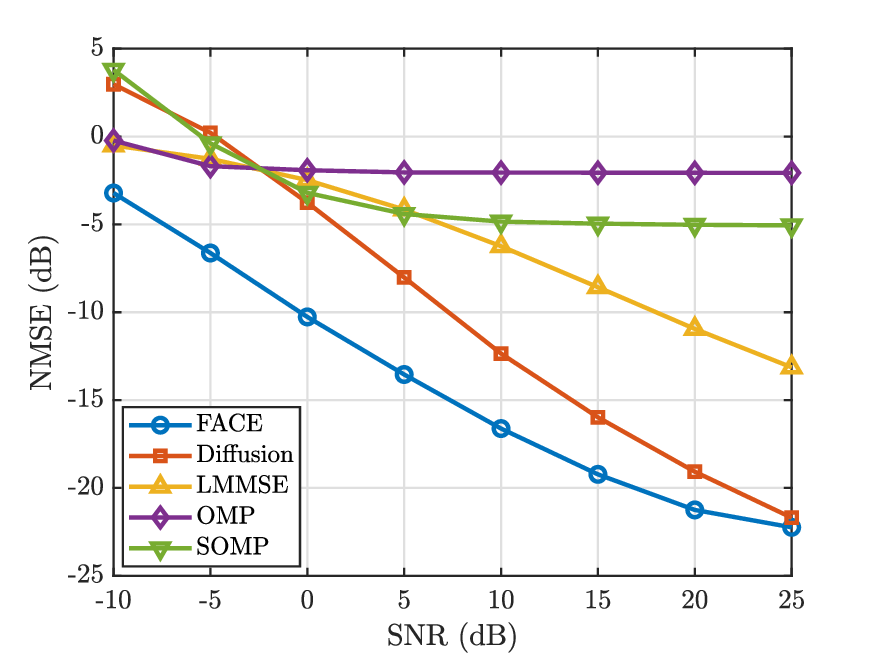}
        \caption{$N_{\textrm{t}} = 32$, $N_{\textrm{r}} = 16$}
        \label{fig:UPA_antennas_32x16}
    \end{subfigure}
    \hfill
    \begin{subfigure}[t]{0.32\textwidth}
        \centering
        \includegraphics[width=1.1\linewidth,trim={5pt 2pt 10pt 10pt},clip]{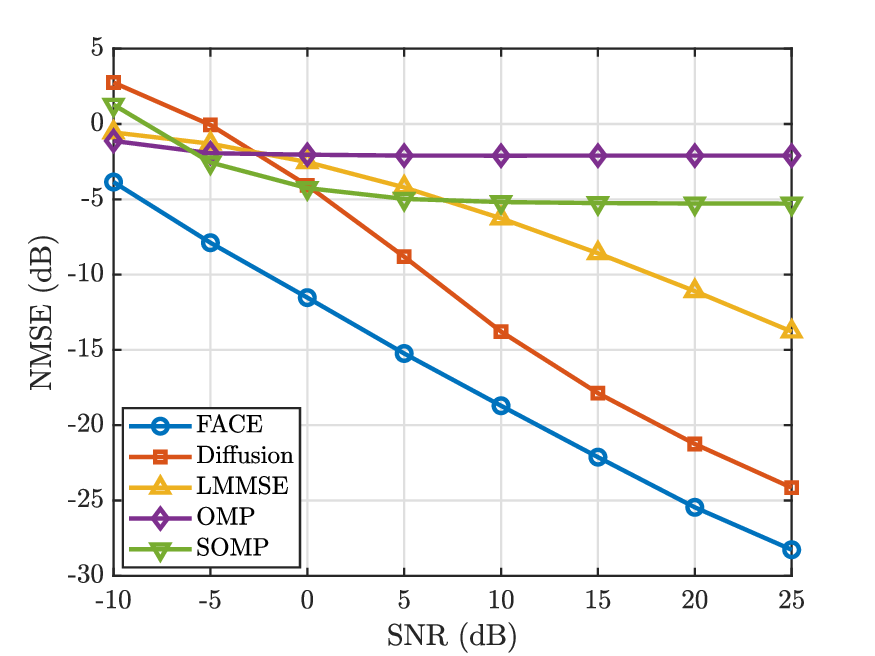}
        \caption{$N_{\textrm{t}} = 64$, $N_{\textrm{r}} = 16$}
        \label{fig:UPA_antennas_64x16}
    \end{subfigure}
    \hfill
    \begin{subfigure}[t]{0.32\textwidth}
        \centering
        \includegraphics[width=1.1\linewidth,trim={5pt 2pt 10pt 10pt},clip]{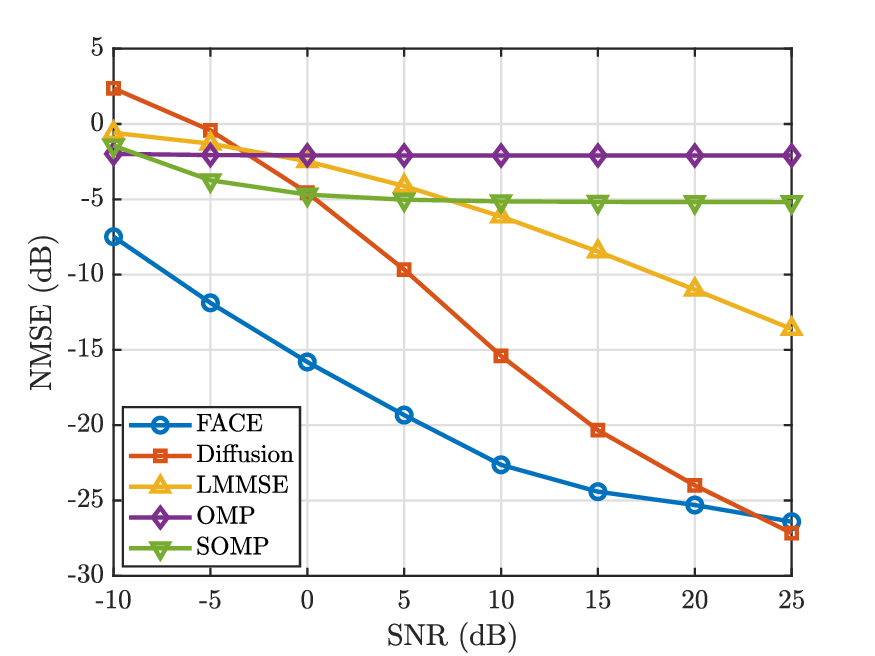}
        \caption{$N_{\textrm{t}} = 128$, $N_{\textrm{r}} = 32$}
        \label{fig:UPA_antennas_128x32}
    \end{subfigure}

    \caption{Variation of \ac{NMSE} with \ac{SNR} for different \ac{UPA} geometry.}
    \label{fig:UPA_antenna_configurations}
\end{figure*}

In the first experiment, we examine the channel estimation performance of \ac{FACE} and the baseline schemes considered under different \ac{ULA} and \ac{UPA} antenna geometries. For the \ac{ULA}, we present the estimation accuracy of $N_{\textrm{t}}=32$ and $N_{\textrm{r}}=16$ configuration in Fig.~\ref{fig:ULA_antennas_32x16}, $N_{\textrm{t}}=64$ and $N_{\textrm{r}}=16$ configuration in Fig.~\ref{fig:ULA_antennas_64x16}, and $N_{\textrm{t}}=128$ and $N_{\textrm{r}}=32$ configuration in Fig.~\ref{fig:ULA_antennas_128x32}. We observe that non-learning baselines (\emph{i.e.}, \ac{LMMSE}, \ac{OMP}, and \ac{SOMP}) show similar trends across the three considered configurations and exhibit high \ac{NMSE}. While \ac{OMP} and \ac{SOMP} demonstrate \ac{NMSE} that is independent of the \ac{SNR} level and the antenna configurations, \ac{LMMSE} gradually enhances the \ac{NMSE} as \ac{SNR} increases, but remains high even at high \ac{SNR} levels. This weak dependence on \ac{SNR} arises from basis mismatch in the predefined quantized angular dictionaries used by \ac{OMP} and \ac{SOMP}, resulting in an error floor that persists as the noise level decreases.

In contrast, the diffusion baseline and the proposed \ac{FACE} framework progressively improve with increasing \ac{SNR} across all three antenna configurations. The diffusion-based estimator exhibits relatively poor accuracy in the low-\ac{SNR} regime and improves considerably as the \ac{SNR} increases; however, it does not outperform \ac{FACE} over the considered \ac{SNR} range. Moreover, \ac{FACE} achieves progressively lower \ac{NMSE} as the number of deployed antennas increases. This improvement indicates that \ac{FACE} effectively exploits the sparse angular structure provided by larger antenna arrays for more accurate channel recovery.

Fig.~\ref{fig:UPA_antenna_configurations} presents the performance for the three \ac{UPA} geometries considered. \Ac{OMP} and \ac{SOMP} exhibit similar trends, with relatively high \ac{NMSE} and weak dependence on \ac{SNR}, whereas \ac{LMMSE} gradually improves as the \ac{SNR} increases. In contrast, both \ac{FACE} and the diffusion-based estimator achieve progressively lower \ac{NMSE} with increasing \ac{SNR} across all three configurations. However, diffusion performs relatively poorly in the low-\ac{SNR} regime. \Ac{FACE} consistently outperforms the considered baselines, except at approximately $25$ dB for $N_{\textrm{t}}=128$ and $N_{\textrm{r}}=32$, where diffusion achieves a slightly lower \ac{NMSE}. Comparing the two array geometries, \ac{FACE} generally achieves higher estimation accuracy in the \ac{ULA} than the \ac{UPA} under the considered configurations.



\begin{figure*}[t!]
    \centering

    \begin{subfigure}[t]{0.32\textwidth}
        \centering
        \includegraphics[width=1.1\linewidth,trim={5pt 2pt 10pt 10pt},clip]{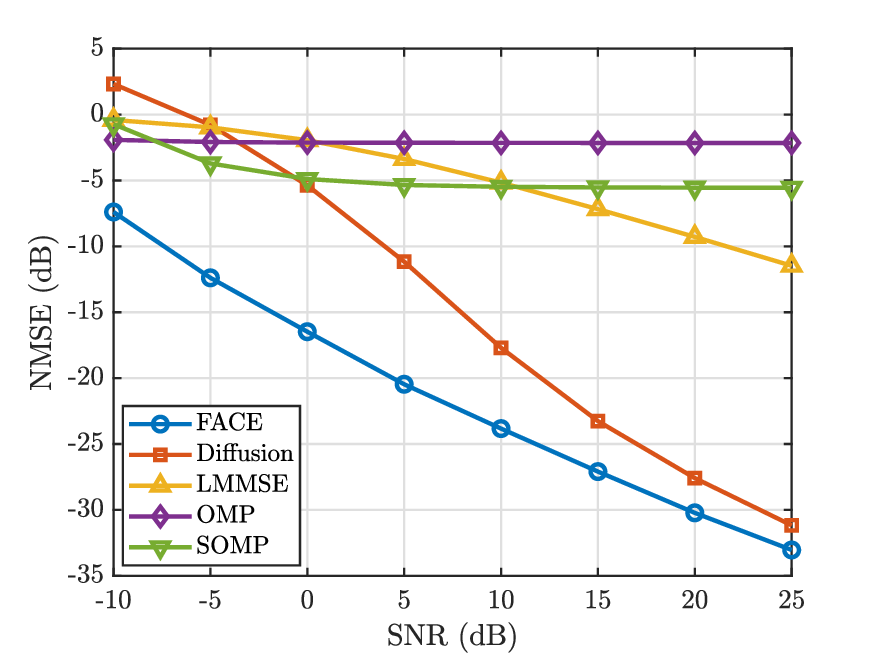}
        \caption{Multi-paths $L = 3$}
        \label{fig:ULA_multipaths_L3}
    \end{subfigure}
    \hfill
    \begin{subfigure}[t]{0.32\textwidth}
        \centering
        \includegraphics[width=1.1\linewidth,trim={5pt 2pt 10pt 10pt},clip]{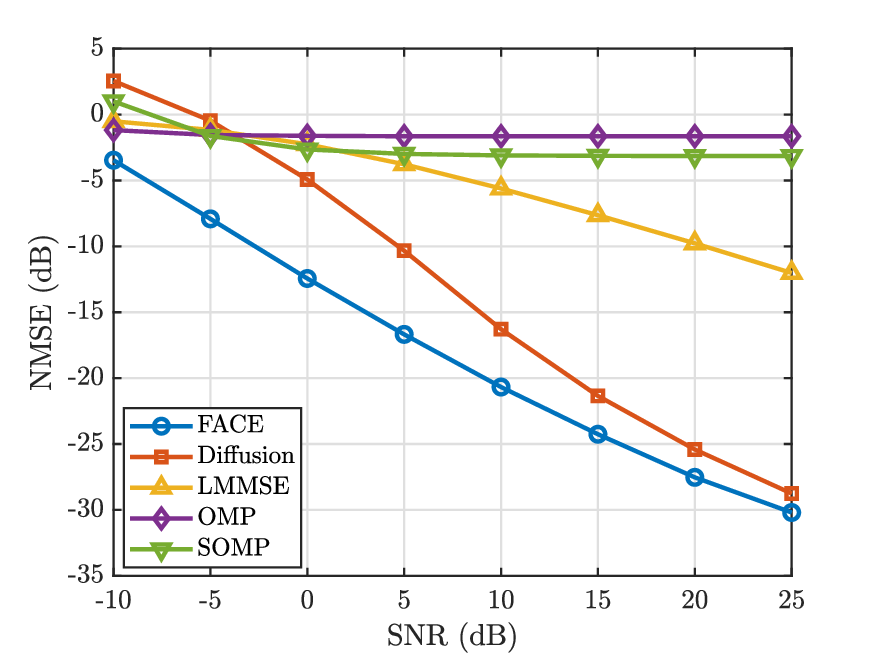}
        \caption{Multi-paths $L = 10$}
        \label{fig:ULA_multipaths_L10}
    \end{subfigure}
    \hfill
    \begin{subfigure}[t]{0.32\textwidth}
        \centering
        \includegraphics[width=1.1\linewidth,trim={5pt 2pt 10pt 10pt},clip]{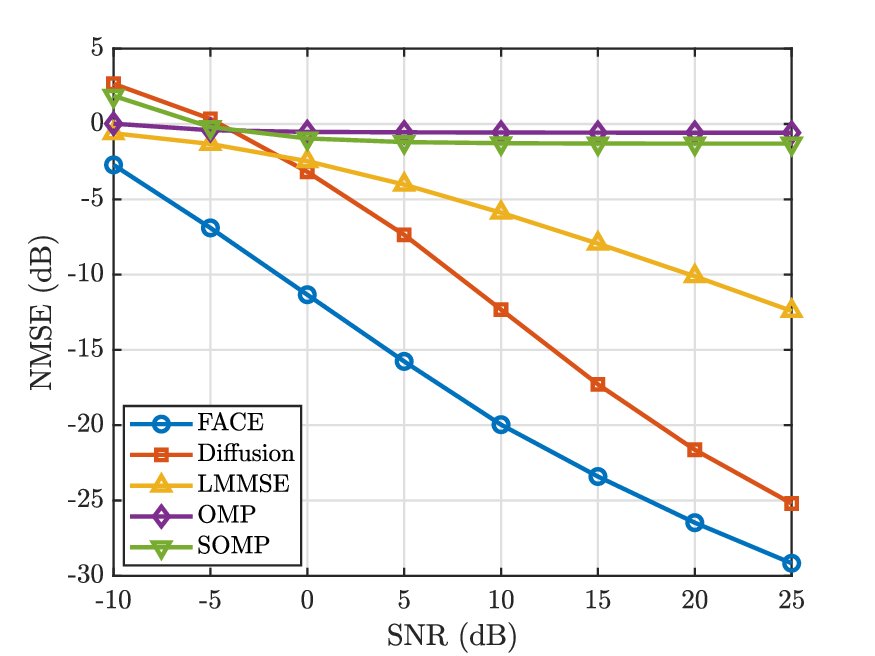}
        \caption{Multi-paths $L = 50$}
        \label{fig:ULA_multipaths_L50}
    \end{subfigure}
    \caption{Variation of \ac{NMSE} with \ac{SNR} for different multi-paths for $N_t = 64$, $N_r = 16$ \ac{ULA} geometry.}
    \vspace{-1mm}
    \label{fig:ULA_multi_paths}
\end{figure*}

\begin{figure*}[t!]
    \centering

    \begin{subfigure}[t]{0.32\textwidth}
        \centering
        \includegraphics[width=1.1\linewidth,trim={5pt 2pt 10pt 10pt},clip]{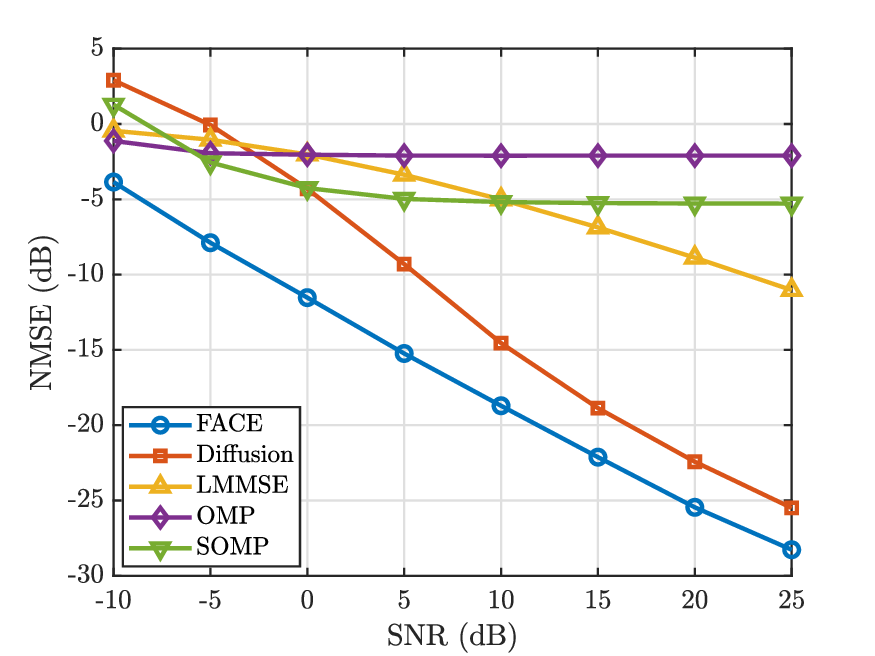}
        \caption{Multi-paths $L = 3$}
        \label{fig:UPA_multipaths_L3}
    \end{subfigure}
    \hfill
    \begin{subfigure}[t]{0.32\textwidth}
        \centering
        \includegraphics[width=1.1\linewidth,trim={5pt 2pt 10pt 10pt},clip]{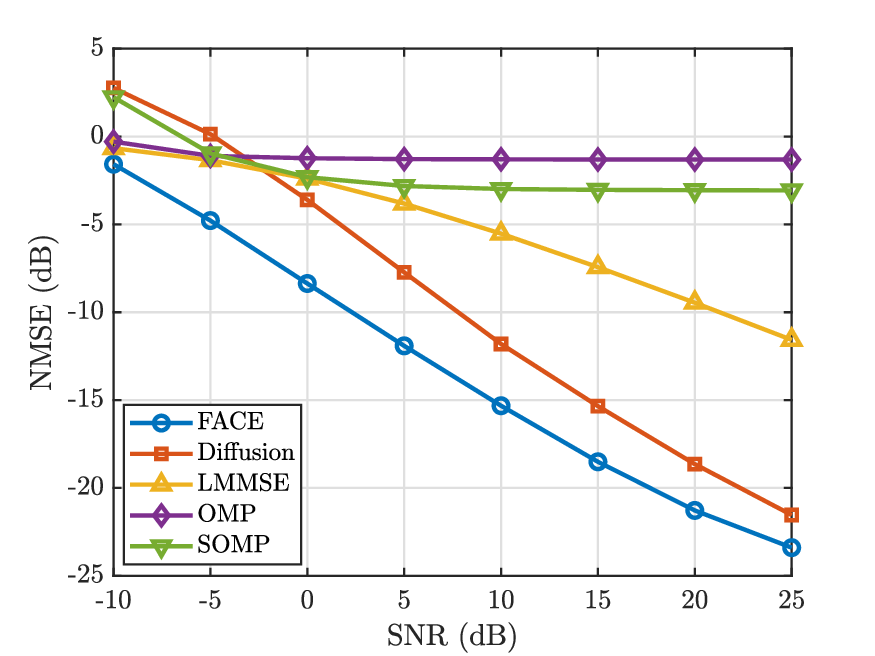}
        \caption{Multi-paths $L = 10$}
        \label{fig:UPA_multipaths_L10}
    \end{subfigure}
    \hfill
    \begin{subfigure}[t]{0.32\textwidth}
        \centering
        \includegraphics[width=1.1\linewidth,trim={5pt 2pt 10pt 10pt},clip]{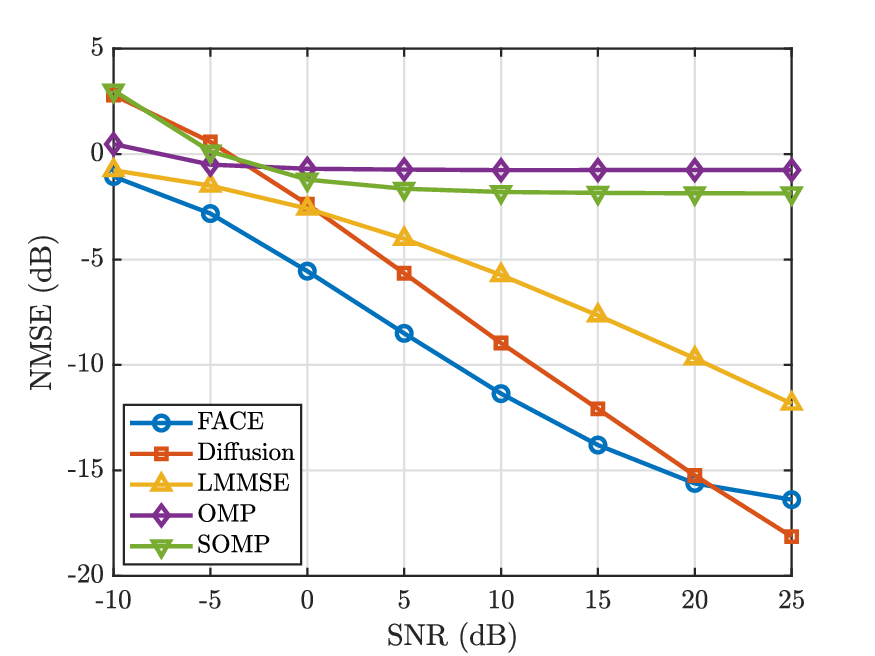}
        \caption{Multi-paths $L = 50$}
        \label{fig:UPA_multipaths_L50}
    \end{subfigure}

    \caption{Variation of \ac{NMSE} with \ac{SNR} for different multi-paths for $N_t = 64$, $N_r = 16$ \ac{UPA} geometry.}
    \vspace{-1mm}
    \label{fig:UPA_multi_paths}
\end{figure*}

In the second experiment, we evaluate the robustness of the considered channel estimation schemes to varying numbers of multipath components for both \ac{ULA} and \ac{UPA} geometries. For the \ac{ULA}, the channel estimation performance for $L=3$, $L=10$, and $L=50$ shown in Fig.~\ref{fig:ULA_multipaths_L3}, Fig.~\ref{fig:ULA_multipaths_L10}, and Fig.~\ref{fig:ULA_multipaths_L50}, respectively. Overall, the considered schemes exhibit similar \ac{NMSE} trends across the three scenarios. However, as the number of multi-path components increases, estimation performance generally degrades, with \ac{OMP} and \ac{SOMP}, exhibiting particularly pronounced degradation. In contrast, \ac{LMMSE} shows a different trend, with its performance remaining relatively stable or slightly improving as $L$ increases.

For the generative approaches, both \ac{FACE} and the diffusion-based estimator exhibit minor performance degradation as the number of multi-path components increases. The diffusion-based estimator performs poorly in the low-\ac{SNR} regime but improves considerably with increasing \ac{SNR}, approaching \ac{FACE} at high \ac{SNR}. In contrast, \ac{FACE} maintains consistently low \ac{NMSE} across the considered \ac{SNR} range and remains comparatively robust to increasing channel richness. As the number of multi-path components increases, more significant angular components emerge, resulting in a less sparse angular-domain channel representation. Despite this increased channel complexity, \ac{FACE} maintains superior estimation accuracy over all considered baseline schemes.

Fig.~\ref{fig:UPA_multipaths_L3}, Fig.~\ref{fig:UPA_multipaths_L10}, and Fig.~\ref{fig:UPA_multipaths_L50} present the channel estimation performance for $L=3$, $L=10$, and $L=50$, respectively, under the \ac{UPA} geometry. Overall, \ac{FACE} and the considered baseline schemes exhibit trends similar to those observed for the \ac{ULA}. Both \ac{FACE} and the diffusion-based estimator achieve progressively lower \ac{NMSE} as the \ac{SNR} increases; however, their overall estimation accuracy degrades as the number of multi-path components increases. \Ac{FACE} outperforms all considered baselines across the three multi-path scenarios, except at approximately $25$ dB for $L=3$, where the diffusion-based estimator achieves slightly lower \ac{NMSE}.



\begin{figure}[t!]
   \centering

   \begin{subfigure}[t]{1\columnwidth}
       \centering
       \includegraphics[width=0.8\linewidth, trim={0 0 0 10pt}, clip]{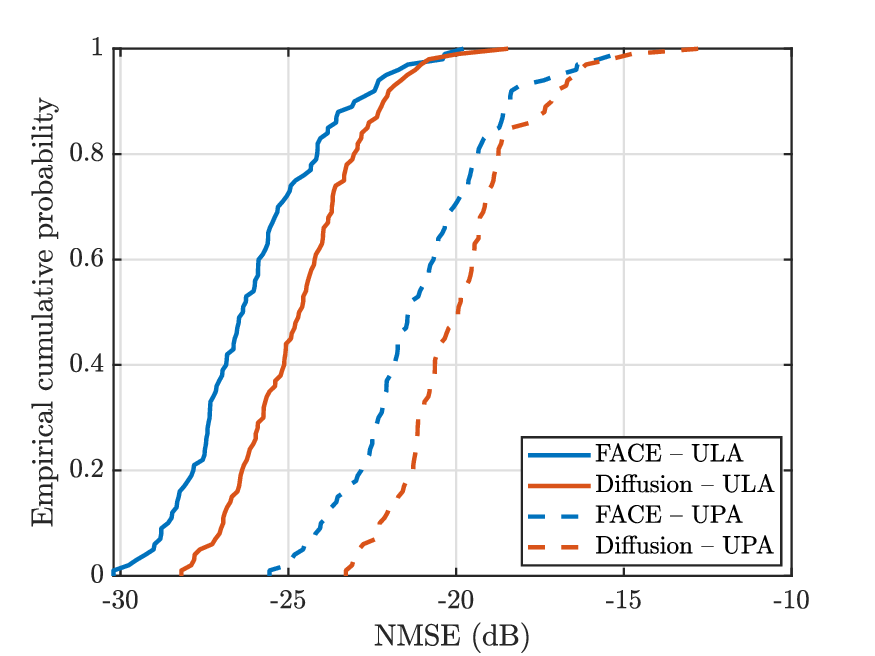}
       \caption{\Ac{CDF} of \ac{ULA} and \ac{UPA} geometry.}
       \label{fig:CDF_ULA_UPA}
   \end{subfigure}
   \hfill
   \begin{subfigure}[t]{1\columnwidth}
       \centering
       \includegraphics[width=0.8\linewidth, trim={0 0 0 10pt}, clip]{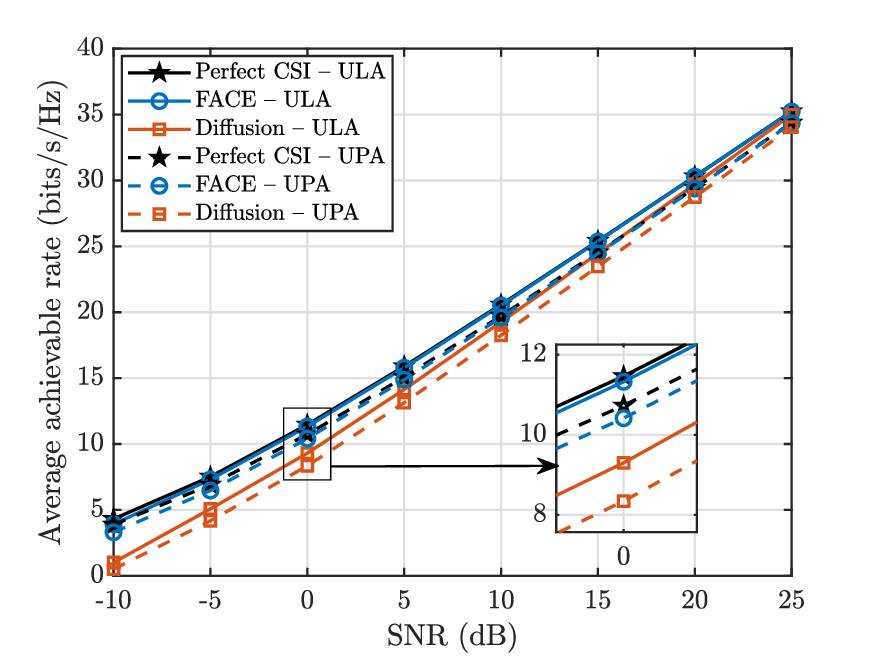}
       \caption{Average achievable rate.}
       \label{fig:ULA_UPA_achievable_rate}
   \end{subfigure}

   \caption{Comparison of the empirical cumulative distribution of the \ac{NMSE} and the average achievable rates of the estimated channels: (a) CDF of \ac{ULA} and \ac{UPA} geometry, and (b) achievable rate of \ac{ULA} and \ac{UPA} geometry.}
   \vspace{-2mm}
   \label{fig:CDF_rate}
\end{figure}

Beyond average estimation performance, it is important to evaluate the consistency of channel estimation accuracy across individual channel realizations. Fig.~\ref{fig:CDF_ULA_UPA} presents the empirical \ac{CDF} of the \ac{NMSE} for \ac{FACE} and the diffusion-based estimator under \ac{ULA} and \ac{UPA} geometries at an \ac{SNR} of $15$ dB. For both geometries, the \ac{CDF} of \ac{FACE} lies to the left of that of the diffusion-based estimator, indicating lower estimation errors across most channel realizations. The similar shapes of the distributions further suggest that the performance advantage of \ac{FACE} is not limited to a few favorable channel realizations. 

Moreover, \ac{FACE} exhibits lower maximum observed \ac{NMSE}, indicating improved estimation reliability across the evaluated samples. Both estimators generally achieve lower \ac{NMSE} under the \ac{ULA} geometry, potentially due to the finer angular resolution along a single dimension in the considered \ac{ULA}, which enables stronger energy concentration in the angular-domain compared with the two-dimensional angular representation of the \ac{UPA}. These results demonstrate that \ac{FACE} maintains its estimation accuracy advantage across both array geometries.

\vspace{0.5mm}

Fig.~\ref{fig:ULA_UPA_achievable_rate} compares the average achievable rate for \ac{ULA} and \ac{UPA} geometries with $N_{\textrm{t}}=64$ and $N_{\textrm{r}}=16$. For both geometries, \ac{FACE} closely tracks the perfect \ac{CSI} benchmark and consistently outperforms the diffusion-based estimator. The \ac{ULA} configuration yields marginally higher achievable rates than the \ac{UPA} across the considered schemes, indicating a modest performance difference associated with the array geometry. This gap narrows at high \ac{SNR}, where the corresponding \ac{ULA} and \ac{UPA} curves converge. Importantly, \ac{FACE} maintains its achievable rate advantage over the diffusion-based estimator for both array geometries, demonstrating consistent performance across the considered antenna structures.


\vspace{-1mm}
\subsection{Results for Real Measurement Channels} \label{sec:real_results}


\begin{figure}[t!]
    \centering

    \begin{subfigure}[t]{1\columnwidth}
        \centering
        \includegraphics[width=0.8\linewidth, trim={0 0 0 10pt}, clip]{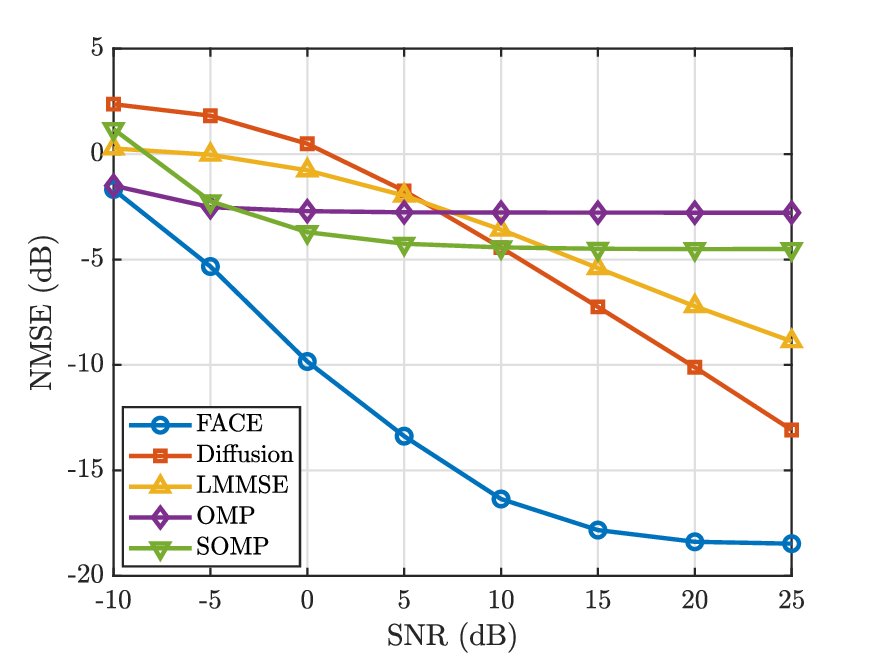}
        \caption{Comparison of \ac{NMSE} with \ac{SNR}.}
        \label{fig:real_nmse}
    \end{subfigure}
    \hfill
    \begin{subfigure}[t]{1\columnwidth}
        \centering
        \includegraphics[width=0.8\linewidth, trim={0 0 0 10pt}, clip]{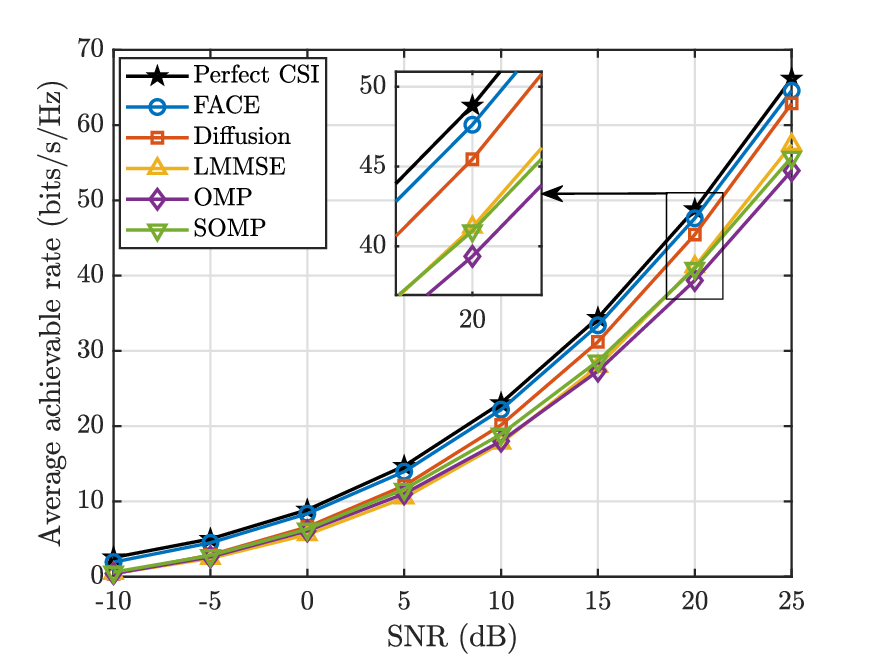}
        \caption{Average achievable rate.}
        \label{fig:real_rate}
    \end{subfigure}

    \caption{Channel estimation performance of the considered estimation
    schemes using real channel measurements: (a) Comparison of \ac{NMSE} with \ac{SNR}
    and (b) average achievable rate.}
    \vspace{-2mm}
    \label{fig:real_data_results}
\end{figure}

Fig.~\ref{fig:real_data_results} presents the channel estimation performance of the considered schemes using measurement-derived channels. Specifically, Fig.~\ref{fig:real_nmse} compares their estimation accuracy across the considered \ac{SNR} range, complementing the preceding synthetic-channel results. \Ac{FACE} consistently achieves the lowest \ac{NMSE} across the entire \ac{SNR} range, demonstrating its effectiveness under realistic propagation conditions. Its estimation accuracy improves substantially at low-to-moderate \ac{SNR}, followed by smaller improvements at high \ac{SNR}. \ac{FACE} maintains a clear performance advantage over both conventional and diffusion-based estimators, although the gap with diffusion narrows as \ac{SNR} increases. Among the conventional schemes, \ac{OMP} and \ac{SOMP} exhibit early \ac{NMSE} saturation, whereas \ac{LMMSE} improves gradually with increasing \ac{SNR}.

Unlike the synthetic-channel results, where the \ac{NMSE} of \ac{FACE} decreases continuously with increasing \ac{SNR}, the measurement-derived results exhibit higher estimation errors and \ac{NMSE} saturation at high \ac{SNR}. This behavior may be attributed to measurement imperfections and uncertainties associated with real-world propagation conditions. Nevertheless, \ac{FACE} consistently outperforms the considered baselines, maintaining a clear performance advantage over the diffusion-based estimator even at high \ac{SNR}, unlike the convergence observed in the synthetic-channel experiments.

Fig.~\ref{fig:real_rate} compares the average achievable rates of the considered schemes using measurement-derived channels, alongside the perfect-\ac{CSI} benchmark. \Ac{FACE} consistently achieves the closest performance to perfect \ac{CSI} across the entire \ac{SNR} range, maintaining only a marginal rate gap even at high \ac{SNR}. This indicates that the accurate channel estimates obtained using \ac{FACE} enable effective transmit precoding, yielding achievable rates close to those obtained with perfect \ac{CSI}. The diffusion-based estimator achieves the second-highest rate. Among the conventional estimators, \ac{LMMSE} and \ac{SOMP} exhibit comparable rates, whereas \ac{OMP} consistently achieves the lowest rate. These observations are consistent with the \ac{NMSE} results in Fig.~\ref{fig:real_nmse}, demonstrating the advantage of \ac{FACE} in translating improved channel estimation accuracy into higher achievable rates.


\vspace{-1mm}
\subsection{Generative Channel Estimation Analysis} \label{sec:genai_analysis}

Thus far, we evaluated \ac{FACE}'s channel estimation performance against baseline schemes by varying key system and channel parameters. Next, we focus on the factors affecting generative channel estimation and therefore restrict subsequent comparisons to \ac{FACE} and the diffusion-based estimator.

\begin{figure}[t!]
    \centering

    \begin{subfigure}[t]{1\columnwidth}
        \centering
        \includegraphics[width=0.8\columnwidth,trim={0 0 0 10pt},clip]{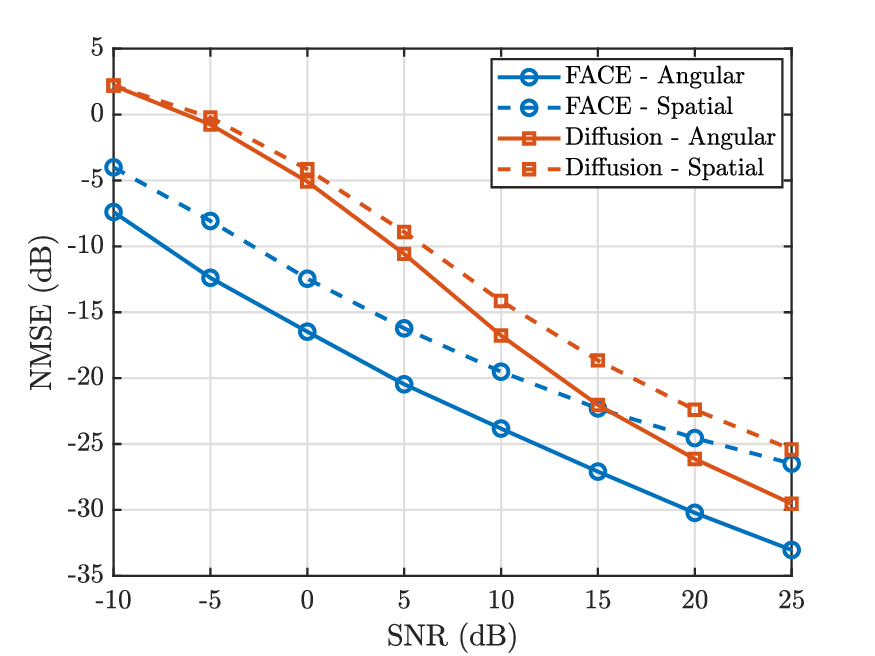}
        \caption{Performance in the spatial and angular domains.}
        \label{fig:domains}
    \end{subfigure}
    \hfill
    \begin{subfigure}[t]{1\columnwidth}
        \centering
        \includegraphics[width=0.8\columnwidth,trim={0 0 0 10pt},clip]{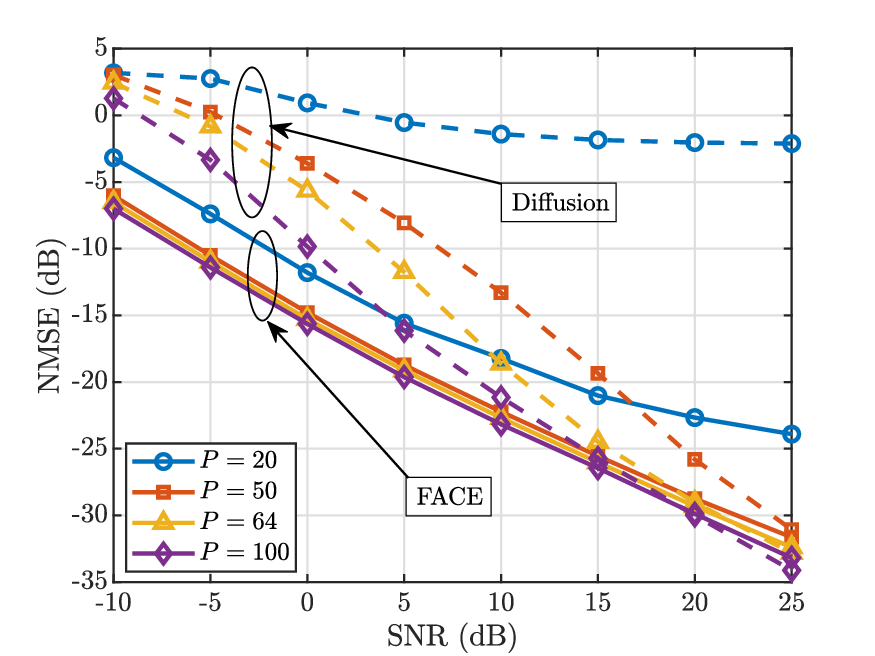}
        \caption{Performance for different pilot lengths.}
        \label{fig:pilot_legths}
    \end{subfigure}

    \caption{Channel estimation performance of \ac{FACE} and the diffusion-based
    estimator: (a) Spatial and angular domains,
    and (b) Pilot length impact.}
    \vspace{-2mm}
    \label{fig:domains_pilot_lengths}
\end{figure}

Fig.~\ref{fig:domains} compares the estimation accuracy of \ac{FACE} and the diffusion-based estimator in the angular and spatial-domains. Both methods achieve consistently lower \ac{NMSE} in the angular-domain across the considered \ac{SNR} range, with \ac{FACE} providing the best overall performance. While the angular-domain gain of the diffusion-based estimator becomes more pronounced at moderate-to-high \ac{SNR}, \ac{FACE} consistently benefits from the angular representation even in the challenging low-\ac{SNR} regime. These results show that exploiting the inherent angular-domain sparsity of \ac{mmWave} channels improves generative channel estimation.


Next, we evaluate the impact of the pilot length on the channel estimation performance of \ac{FACE} and the diffusion-based estimator in the $N_{\textrm{t}}=64$ configuration as shown in Fig.~\ref{fig:pilot_legths}. We vary the pilot length $P\in \{20,50,64,100\}$, where $P=20$ and $P=50$ correspond to undersampled pilot settings, while $P=100$ represents an oversampled setting. \ac{FACE} remains closely clustered across the considered \ac{SNR} range, demonstrating limited sensitivity to pilot reduction. At the shortest pilot length, $P=20$, channel estimation accuracy degrades only modestly compared with larger pilot settings.

In contrast, the diffusion-based estimator exhibits a higher sensitivity to pilot reduction, with a substantial performance gap among the corresponding \ac{NMSE} curves. At $P=20$, its \ac{NMSE} saturates early and shows negligible improvement as \ac{SNR} increases. However, at $P \geq 50$, the estimation accuracy improves considerably and progressively approaches that of \ac{FACE} at higher \ac{SNR}. These results highlight the superior pilot efficiency of \ac{FACE}, which retains most of its estimation accuracy even when the pilot length is reduced by approximately $3\times$ relative to $N_{\textrm{t}}$. 
In undersampled settings, the diffusion-based estimator incorporates measurements through guidance of an unconditional reverse process, potentially limiting measurement consistency. In contrast, \ac{FACE} explicitly promotes measurement consistency at each \ac{PnP-PGD} iteration while leveraging the learned flow matching prior to recover unobserved channel components.

\begin{figure}[t!]
    \centering
    \begin{subfigure}{1\linewidth}
        \centering
        \includegraphics[width=1.05\columnwidth,trim={10 0 10 0pt},clip]{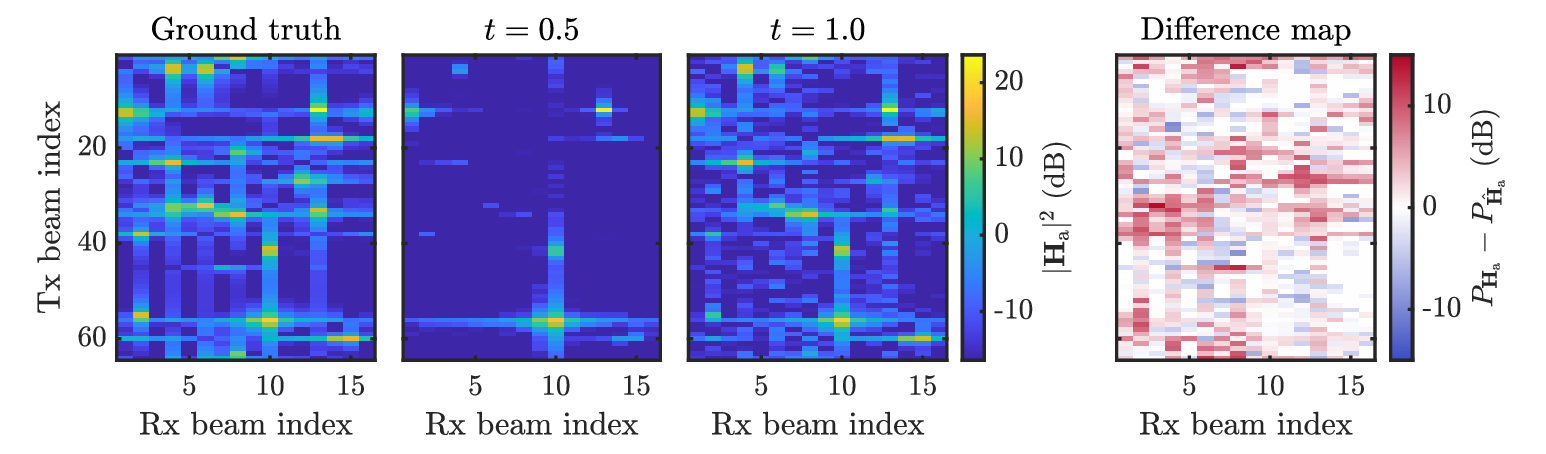}
        \caption{\ac{FACE} evolution over time.}
        \label{subfig:fm_evolution}
    \end{subfigure}
    
    \begin{subfigure}{1\linewidth}
        \centering
        \includegraphics[width=1.05\columnwidth,trim={10 0 10 0pt},clip]{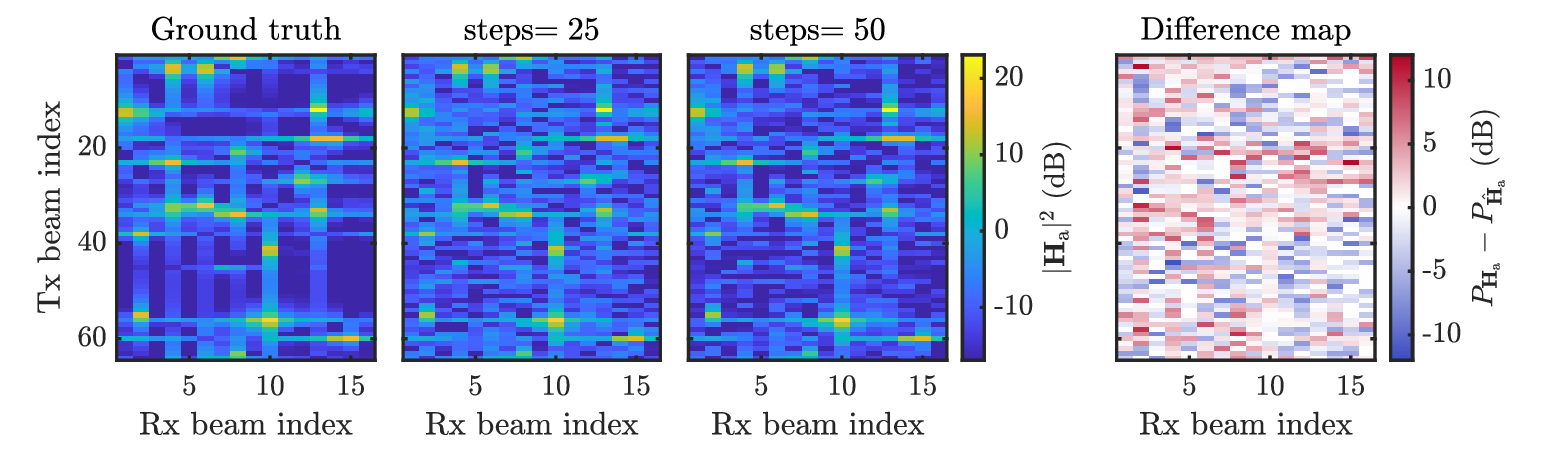}
        \caption{Diffusion evolution over steps.}
        \label{subfig:diff_evolution}
    \end{subfigure}

    \caption{The channel estimation evolution and difference map for multi-paths $L = 20$: (a) \ac{FACE} evolution over time, (b) diffusion evolution over steps.}
    \vspace{-2mm}
    \label{fig:evolution}
\end{figure}

Fig.~\ref{fig:evolution} illustrates the evolution of the reconstructed angular-domain channel energy during inference for \ac{FACE} and the diffusion-based estimator, along with their corresponding difference maps relative to the ground truth. Starting from a noise-like initialization, \ac{FACE} progressively recovers the sparse angular structure, with dominant components emerging at $t=0.5$ and weaker components becoming distinguishable by $t=1.0$. The resulting reconstruction closely resembles the ground truth energy distribution. In contrast, the diffusion-based estimator exhibits residual energy across numerous angular bins at steps $25$ and $50$, resulting in a less distinct sparse structure. The difference maps further highlight the superior reconstruction accuracy of \ac{FACE}, showing negligible deviations across most angular bins at $t=1.0$, with minor errors concentrated around the dominant components. Meanwhile, the diffusion-based estimator exhibits more widely distributed reconstruction errors at step $50$, indicating less accurate recovery of the angular-domain channel structure.

\begin{figure}[t!]
    \centering

    \begin{subfigure}[t]{1\columnwidth}
        \centering
        \includegraphics[width=0.8\columnwidth,trim={0 0 0 10pt},clip]{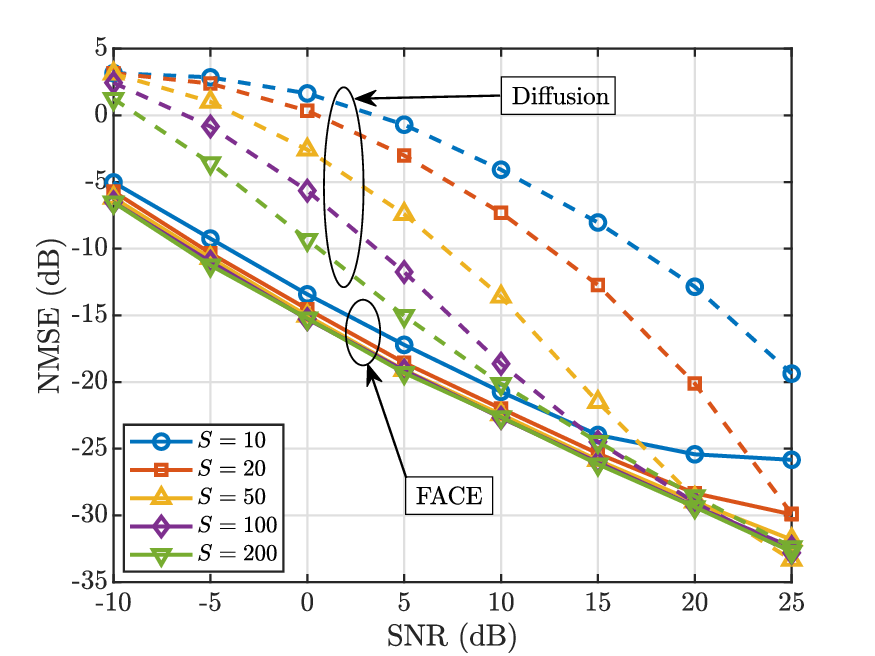}
        \caption{Estimation performance for different numbers of sampling steps.}
        \label{fig:step_sizes}
    \end{subfigure}
    \hfill
    \begin{subfigure}[t]{1\columnwidth}
        \centering
        \includegraphics[width=0.8\columnwidth,trim={0 0 0 10pt},clip]{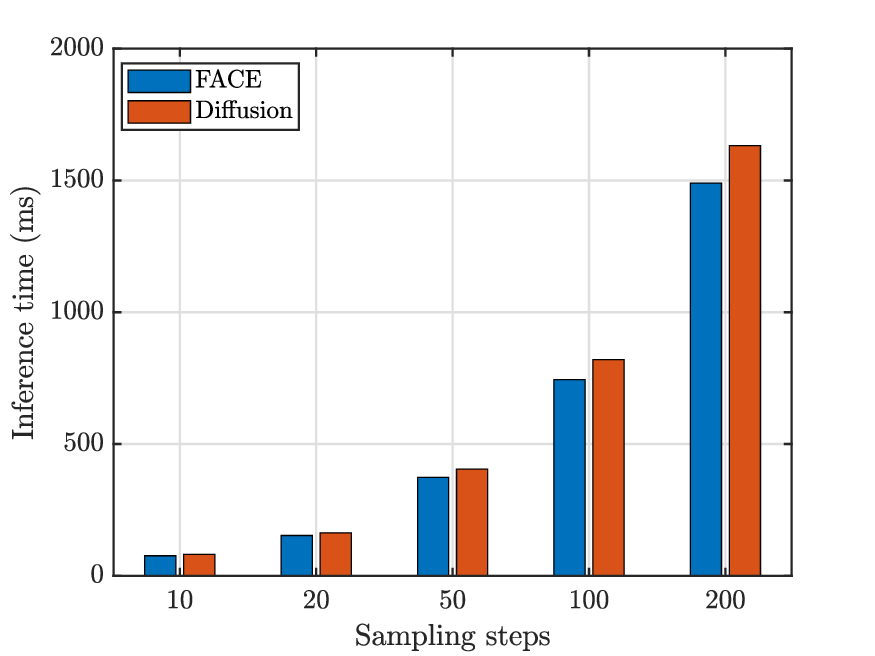}
        \caption{Inference time for different numbers of sampling steps.}
        \label{fig:inference_time}
    \end{subfigure}

    \caption{Impact of the number of sampling steps on \ac{FACE} and the diffusion-based estimator: (a) Channel estimation performance and
    (b) inference time.}
    \label{fig:inferene_time_step_size}
\end{figure}

Sensitivity to the number of sampling steps is an important consideration for generative channel estimation. Fig.~\ref{fig:step_sizes} compares the performance of \ac{FACE} and the diffusion-based estimator for different step sizes $S\in \{10,20,50,100,200\}$. The two methods exhibit markedly different sensitivities to $S$. The \ac{NMSE} curves of \ac{FACE} remain nearly indistinguishable across the considered \ac{SNR} range, with only a slight degradation at high \ac{SNR} for $S=10$. In contrast, the diffusion-based estimator exhibits substantial performance variation with $S$, showing poor estimation accuracy for $S=10$ and significantly higher \ac{NMSE} across the entire considered \ac{SNR} range. Its estimation accuracy progressively improves as $S$ increases, approaching that of \ac{FACE} in the high-\ac{SNR} regime for $S \geq 50$.

This difference stems from the underlying generative sampling processes. Flow matching promotes near-straight transport trajectories between the source and target distributions, simplifying the numerical integration of the learned velocity field and enabling accurate channel estimation with relatively few sampling steps~\cite{flow_based_generative_models}. In contrast, diffusion-based sampling follows a more complex reverse trajectory governed by the learned reverse-time dynamics and noise schedule, typically requiring finer discretization. Therefore, too few sampling steps introduce larger discretization errors, causing deviations from the desired reverse trajectory and degrading channel estimation accuracy.

Fig.~\ref{fig:inference_time} compares the inference time of \ac{FACE} and the diffusion-based estimator for different numbers of sampling steps. For both methods, the inference time increases approximately linearly with the number of sampling steps, as each additional step requires a separate network evaluation. The diffusion-based estimator exhibits slightly higher inference time than \ac{FACE} across all considered sampling steps. These empirical results are consistent with the computational complexity analysis presented next in Section~\ref{sec: complexity}, further highlighting the inference efficiency of the proposed \ac{FACE} framework.



\subsection{Complexity Analysis} \label{sec: complexity}

Since \ac{FACE} and the diffusion baseline share an identical U-Net backbone, a single forward pass incurs the same computational cost for both methods, denoted by $C_{\textrm{net}}$. For a convolutional layer $l$ with kernel size $k_{l}^{\textrm{c}} \times k_{l}^{\textrm{c}}$, $C_{l}^{\textrm{in}}$ input channels, $C_{l}^{\textrm{out}}$ output channels, and output dimensions $H_{l}^{\textrm{out}} \times W_{l}^{\textrm{out}}$, the computational complexity is approximately $\mathcal{O}\big((k_{l}^{\textrm{c}})^2 H_{l}^{\textrm{out}} W_{l}^{\textrm{out}} C_{l}^{\textrm{in}} C_{l}^{\textrm{out}} \big)$.
Accordingly, summing the computational cost over the convolutional operations across the U-Net resolution levels is given by:
\begin{equation}
C_{\textrm{net}} = \mathcal{O}\left(\sum_{l=1}^{L_{\textrm{r}}} (k_{l}^{\textrm{c}})^2 H_{l}^{\textrm{out}}W_{l}^{\textrm{out}}C_{l}^{\textrm{in}}C_{l}^{\textrm{out}} \right),
\end{equation}
where $L_{\textrm{r}}$ denotes the number of considered U-Net resolution levels.

At each \ac{PnP-PGD} iteration, \ac{FACE} performs one forward evaluation of the flow matching denoiser together with a data-fidelity update, resulting in a per-iteration complexity of $C_{\textrm{net}}+\mathcal{O}(N_{\textrm{r}}N_{\textrm{t}}P)$. Similarly, each reverse sampling step of the diffusion-based estimator requires one U-Net evaluation and an \ac{SVD}-guided measurement-consistency update with complexity $\mathcal{O}(N_{\textrm{r}}N_{\textrm{t}}P)$. In addition, the diffusion-based estimator requires a one-time \ac{SVD} computation, with complexity $\mathcal{O}\big(\min(N_{\textrm{r}},N_{\textrm{t}})^2\max(N_{\textrm{r}},N_{\textrm{t}})\big)$. Accordingly, the overall inference complexity of \ac{FACE} is: 
\begin{equation} \label{eq:FACE_complexity}
    C_{\textrm{FACE}} = \mathcal{O}\big(K_{\textrm{f}} (C_{\textrm{net}} + N_{\textrm{r}}N_{\textrm{t}}P)\big), 
\end{equation}
where $K_{\textrm{f}}$ denotes the number of sampling steps used by \ac{FACE}. In comparison, the overall complexity of the diffusion-based estimator is:
\begin{align} \label{eq:DIFF_complexity}
    C_{\textrm{Diff}} = \: &\mathcal{O}\big(K_{\textrm{d}}(C_{\textrm{net}}
    + N_{\textrm{r}}N_{\textrm{t}}P)\big) + \nonumber\\
\quad &\mathcal{O}\big( \min(N_{\textrm{r}},N_{\textrm{t}})^2 \max(N_{\textrm{r}},N_{\textrm{t}}) \big), 
\end{align}
where $K_{\textrm{d}}$ denotes the number of reverse sampling steps used by the diffusion-based estimator. For the $N_{\textrm{t}}=64$ and $N_{\textrm{r}}=16$ antenna configuration, both \ac{FACE} and the diffusion-based estimator require approximately $1.025$ \ac{GFLOPs} per sampling step. Although the computational cost of a single network evaluation is therefore nearly identical, \ac{FACE} requires substantially fewer sampling steps to achieve accurate channel estimation, resulting in a lower overall inference cost than the diffusion-based estimator. Table~\ref{table:Complexity} compares the computational costs of \ac{FACE} and the diffusion-based estimator, measured in \ac{GFLOPs}, required to achieve a target \ac{NMSE} for the considered antenna configuration.

\begin{table}[t!] 
\centering
\caption{Computational cost comparison of \ac{FACE} and the diffusion-based estimator in terms of \ac{GFLOPs} for achieving a target \ac{NMSE}}
\label{table:Complexity}
\begin{tabular}{@{}ccc@{}}
\toprule
\textbf{Approximate \ac{NMSE}} \quad & \textbf{\ac{FACE} complexity} & \textbf{Diffusion complexity}\\ 
\midrule
$-20$ dB & $10.25$ & $205.00$\\
$-25$ dB & $20.50$ & $205.00$ \\
\bottomrule
\end{tabular}
\end{table}

\section{Conclusion} \label{sec: Conclusion}

In this work, we proposed an efficient flow matching-based framework for \ac{mmWave} \ac{MIMO} channel estimation by exploiting the inherent sparsity of channels in the angular-domain. We first formulated channel estimation as a \ac{MAP}-based regularized inverse problem that combines measurement-driven data fidelity with a learned channel prior. Building on this formulation, we developed \ac{FACE}, a novel channel estimation framework that integrates a flow matching-based generative prior with a \ac{PnP-PGD} solver for angular-domain channel recovery. Extensive numerical results demonstrated that \ac{FACE} consistently achieves superior estimation accuracy compared with \ac{LMMSE}, \ac{OMP}, \ac{SOMP}, and the diffusion-based estimator across a wide range of \ac{SNR}, multi-path, and pilot configurations. Moreover, \ac{FACE} achieves an average achievable rate close to the perfect \ac{CSI} benchmark while maintaining high estimation accuracy under reduced pilot overhead. Importantly, \ac{FACE} requires substantially fewer sampling steps than the diffusion-based estimator to achieve comparable accuracy, resulting in lower inference time and computational cost. These results establish \ac{FACE} as an accurate and computationally efficient approach for channel estimation in resource-constrained \ac{mmWave} \ac{MIMO} systems. Extending the proposed \ac{FACE} framework for interference management in \ac{mmWave} \ac{MIMO} systems is left for future work.

\bibliographystyle{IEEEtran}
\bibliography{references}

@ARTICLE{Diff_based_MIMO_BF,
  author={Liu, Haoqiang and Chen, Huiming and Huang, Wenzhen and Wei, Zhaobin and Zeng, Yonghong and Yan, Hong},
  journal={IEEE Trans. Wireless Commun.}, 
  title={Coordinated Downlink Beamforming in Multi-Cell {MIMO} Networks: A Diffusion Model-Enhanced Multi-Agent Reinforcement Learning Perspective}, 
  year={2026},
  volume={25},
  number={},
  pages={7617-7634},
  doi={10.1109/TWC.2025.3632855}}

@ARTICLE{DiffPace,
  author={Hu, Zhengdong and Han, Chong and Gerstacker, Wolfgang and Schober, Robert},
  journal={IEEE J. Sel. Areas Commun.}, 
  title={DiffPace: Diffusion-Based Plug-and-Play Augmented Channel Estimation in {mmWave} and Terahertz Ultra-Massive {MIMO} Systems}, 
  year={2026},
  volume={44},
  number={},
  pages={2559-2573},
  doi={10.1109/JSAC.2025.3643404}}

@ARTICLE{mmWave_THz_com,
  author={Akyildiz, Ian F. and Han, Chong and Nie, Shuai},
  journal={IEEE Commun. Mag.}, 
  title={Combating the Distance Problem in the Millimeter Wave and Terahertz Frequency Bands}, 
  year={2018},
  volume={56},
  number={6},
  pages={102-108},
  doi={10.1109/MCOM.2018.1700928}}

@ARTICLE{CE_using_CS,
  author={Alkhateeb, Ahmed and El Ayach, Omar and Leus, Geert and Heath, Robert W.},
  journal={IEEE J. Sel. Topics Signal Process.}, 
  title={Channel Estimation and Hybrid Precoding for Millimeter Wave Cellular Systems}, 
  year={2014},
  volume={8},
  number={5},
  pages={831-846},
  doi={10.1109/JSTSP.2014.2334278}}

@ARTICLE{CE_OMP,
  author={Lee, Junho and Gil, Gye-Tae and Lee, Yong H.},
  journal={IEEE Trans. Commun.}, 
  title={Channel Estimation via Orthogonal Matching Pursuit for Hybrid {MIMO} Systems in Millimeter Wave Communications}, 
  year={2016},
  volume={64},
  number={6},
  pages={2370-2386},
  doi={10.1109/TCOMM.2016.2557791}}

@ARTICLE{CE_LMMSE,
  author={Nayebi, Elina and Rao, Bhaskar D.},
  journal={IEEE Trans. Signal Process.}, 
  title={Semi-blind Channel Estimation for Multiuser Massive {MIMO} Systems}, 
  year={2018},
  volume={66},
  number={2},
  pages={540-553},
  doi={10.1109/TSP.2017.2771725}}

@ARTICLE{CE_joint_sparse_low_rank,
  author={Li, Xingjian and Fang, Jun and Li, Hongbin and Wang, Pu},
  journal={IEEE Trans. Wireless Commun.}, 
  title={Millimeter Wave Channel Estimation via Exploiting Joint Sparse and Low-Rank Structures}, 
  year={2018},
  volume={17},
  number={2},
  pages={1123-1133},
  doi={10.1109/TWC.2017.2776108}}

@ARTICLE{CE_using_LSE,
  author={Huang, Chongwen and Liu, Lei and Yuen, Chau and Sun, Sumei},
  journal={IEEE Trans. Signal Process.}, 
  title={Iterative Channel Estimation Using {LSE} and Sparse Message Passing for MmWave {MIMO} Systems}, 
  year={2019},
  volume={67},
  number={1},
  pages={245-259},
  doi={10.1109/TSP.2018.2879620}}

@ARTICLE{near_field_CE_for_XL_MIMO,
  author={Lu, Yu and Dai, Linglong},
  journal={IEEE Trans. Commun.}, 
  title={Near-Field Channel Estimation in Mixed {LoS/NLoS} Environments for Extremely Large-Scale {MIMO} Systems}, 
  year={2023},
  volume={71},
  number={6},
  pages={3694-3707},
  doi={10.1109/TCOMM.2023.3260242}}

@ARTICLE{accurate_ce_mmWave,
  author={Cheng, Xiantao and Tang, Chao and Zhang, Zhongpei},
  journal={IEEE Trans. Veh. Technol.}, 
  title={Accurate Channel Estimation for Millimeter-Wave {MIMO} Systems}, 
  year={2019},
  volume={68},
  number={5},
  pages={5159-5163},
  doi={10.1109/TVT.2019.2905640}}

@ARTICLE{overview_of_mmWave_sigpro,
  author={Heath, Robert W. and González-Prelcic, Nuria and Rangan, Sundeep and Roh, Wonil and Sayeed, Akbar M.},
  journal={IEEE J. Sel. Topics Signal Process.}, 
  title={An Overview of Signal Processing Techniques for Millimeter Wave {MIMO} Systems}, 
  year={2016},
  volume={10},
  number={3},
  pages={436-453},
  doi={10.1109/JSTSP.2016.2523924}}

@ARTICLE{small_scale_MIMO_CE_1,
  author={Bjornson, Emil and Ottersten, Björn},
  journal={IEEE Trans. Signal Process.}, 
  title={A Framework for Training-Based Estimation in Arbitrarily Correlated Rician {MIMO} Channels With Rician Disturbance}, 
  year={2010},
  volume={58},
  number={3},
  pages={1807-1820},
  doi={10.1109/TSP.2009.2037352}}

@ARTICLE{small_scale_MIMO_CE_2,
  author={Hassibi, B. and Hochwald, B.M.},
  journal={IEEE Trans. Inf. Theory}, 
  title={How much training is needed in multiple-antenna wireless links?}, 
  year={2003},
  volume={49},
  number={4},
  pages={951-963},
  doi={10.1109/TIT.2003.809594}}

@ARTICLE{DL_based_CE,
  author={Gao, Jiabao and Zhong, Caijun and Li, Geoffrey Ye and Soriaga, Joseph B. and Behboodi, Arash},
  journal={IEEE Trans. Commun.}, 
  title={Deep Learning-Based Channel Estimation for Wideband Hybrid MmWave Massive {MIMO}}, 
  year={2023},
  volume={71},
  number={6},
  pages={3679-3693},
  doi={10.1109/TCOMM.2023.3258484}}

@ARTICLE{CE_beamspace_ce_using_DL,
  author={Wei, Xiuhong and Hu, Chen and Dai, Linglong},
  journal={IEEE Trans. Commun.}, 
  title={Deep Learning for Beamspace Channel Estimation in Millimeter-Wave Massive {MIMO} Systems}, 
  year={2021},
  volume={69},
  number={1},
  pages={182-193},
  doi={10.1109/TCOMM.2020.3027027}}

@ARTICLE{CE_in_mMIMO_using_untrained_DNN,
  author={Balevi, Eren and Doshi, Akash and Andrews, Jeffrey G.},
  journal={IEEE Trans. Wireless Commun.}, 
  title={Massive {MIMO} Channel Estimation With an Untrained Deep Neural Network}, 
  year={2020},
  volume={19},
  number={3},
  pages={2079-2090},
  doi={10.1109/TWC.2019.2962474}}

@ARTICLE{CE_gen_diff_models_for_ce,
  author={Zhou, Xingyu and Liang, Le and Zhang, Jing and Jiang, Peiwen and Li, Yong and Jin, Shi},
  journal={IEEE Trans. Wireless Commun.}, 
  title={Generative Diffusion Models for High Dimensional Channel Estimation}, 
  year={2025},
  volume={24},
  number={7},
  pages={5840-5854},
  doi={10.1109/TWC.2025.3549592}}

@ARTICLE{CE_deep_gen_models,
  author={Jin, Zhenzhou and You, Li and Wing Kwan Ng, Derrick and Xia, Xiang-Gen and Gao, Xiqi},
  journal={IEEE Trans. Cogn. Commun. Netw.}, 
  title={Near-Field Channel Estimation for {XL-MIMO}: A Deep Generative Model Guided by Side Information}, 
  year={2026},
  volume={12},
  number={},
  pages={628-643},
  doi={10.1109/TCCN.2025.3566047}}

@ARTICLE{GenAI_survey,
  author={Van Huynh, Nguyen and Wang, Jiacheng and Du, Hongyang and Hoang, Dinh Thai and Niyato, Dusit and Nguyen, Diep N. and Kim, Dong In and Letaief, Khaled B.},
  journal={IEEE Trans. Cogn. Commun. Netw.}, 
  title={Generative {AI} for Physical Layer Communications: A Survey}, 
  year={2024},
  volume={10},
  number={3},
  pages={706-728},
  doi={10.1109/TCCN.2024.3384500}}

@ARTICLE{CE_Conditional_GANs,
  author={Ye, Hao and Liang, Le and Li, Geoffrey Ye and Juang, Biing-Hwang},
  journal={IEEE Trans. Wireless Commun.}, 
  title={Deep Learning-Based End-to-End Wireless Communication Systems With Conditional {GANs} as Unknown Channels}, 
  year={2020},
  volume={19},
  number={5},
  pages={3133-3143},
  doi={10.1109/TWC.2020.2970707}}

@article{Denoising_diff_prob_models,
  title={Denoising diffusion probabilistic models},
  author={Ho, Jonathan and Jain, Ajay and Abbeel, Pieter},
  journal={Advances in neural information processing systems},
  volume={33},
  pages={6840--6851},
  year={2020}
}

@article{flow_based_generative_models,
  title     = {Improving and generalizing flow-based generative models with minibatch optimal transport},
  author    = {Tong, Alexander and Fatras, Kilian and Malkin, Nikolay and Huguet, Guillaume and Zhang, Yanlei and Rector-Brooks, Jarrid and Wolf, Guy and Bengio, Yoshua},
  journal   = {Transactions on Machine Learning Research},
  year      = {2024},
  issn      = {2835-8856},
  url       = {https://openreview.net/forum?id=CD9Snc73AW},
  note      = {Expert Certification}
}

@ARTICLE{sig_pro_ill_posed_claim,
  author={Candes, Emmanuel J. and Wakin, Michael B.},
  journal={IEEE Signal Process. Mag.}, 
  title={An Introduction To Compressive Sampling}, 
  year={2008},
  volume={25},
  number={2},
  pages={21-30},
  doi={10.1109/MSP.2007.914731}}

@ARTICLE{basis_mismatch_CS,
  author={Chi, Yuejie and Scharf, Louis L. and Pezeshki, Ali and Calderbank, A. Robert},
  journal={IEEE Trans. Signal Process.}, 
  title={Sensitivity to Basis Mismatch in Compressed Sensing}, 
  year={2011},
  volume={59},
  number={5},
  pages={2182-2195},
  doi={10.1109/TSP.2011.2112650}}

@ARTICLE{DL_estimator,
  author={Soltani, Mehran and Pourahmadi, Vahid and Mirzaei, Ali and Sheikhzadeh, Hamid},
  journal={IEEE Commun. Lett.}, 
  title={Deep Learning-Based Channel Estimation}, 
  year={2019},
  volume={23},
  number={4},
  pages={652-655},
  doi={10.1109/LCOMM.2019.2898944}}

@ARTICLE{score_based_models,
  author={Arvinte, Marius and Tamir, Jonathan I.},
  journal={IEEE Trans. Wireless Commun.}, 
  title={MIMO Channel Estimation Using Score-Based Generative Models}, 
  year={2023},
  volume={22},
  number={6},
  pages={3698-3713},
  doi={10.1109/TWC.2022.3220784}}

@ARTICLE{CS_techniques,
  author={Jang, Suhwan and Lee, Chungyong},
  journal={IEEE Trans. Wireless Commun.}, 
  title={Learning-Aided Channel Estimation for Wideband {mmWave MIMO} Systems With Beam Squint}, 
  year={2025},
  volume={24},
  number={1},
  pages={706-720},
  doi={10.1109/TWC.2024.3498317}}

@ARTICLE{mmWave_pros,
  author={Rangan, Sundeep and Rappaport, Theodore S. and Erkip, Elza},
  journal={Proc. IEEE}, 
  title={Millimeter-Wave Cellular Wireless Networks: Potentials and Challenges}, 
  year={2014},
  volume={102},
  number={3},
  pages={366-385},
  doi={10.1109/JPROC.2014.2299397}}

@inproceedings{FM_original,
  title={Flow Matching for Generative Modeling},
  author={Yaron Lipman and Ricky T. Q. Chen and Heli Ben-Hamu and Maximilian Nickel and Matt Le},
  booktitle={International Conference on Learning Representations},
  year={2023},
  url={https://openreview.net}
}

@inproceedings{Unet,
  title={{U-net}: Convolutional networks for biomedical image segmentation},
  author={Ronneberger, Olaf and Fischer, Philipp and Brox, Thomas},
  booktitle={International Conference on Medical image computing and computer-assisted intervention},
  pages={234--241},
  year={2015},
  organization={Springer}
}

@INPROCEEDINGS{FM_CE_ICC_paper,
  author={Zhang, Yongqiang and Nadeem, Qurrat-Ul-Ain},
  booktitle={ICC 2026 - IEEE International Conference on Communications}, 
  title={Channel Estimation in {MIMO} Systems Using Flow Matching Models}, 
  year={2026},
  volume={},
  number={},
  pages={1-6},
  doi={10.1109/ICC59461.2026.11587599}}

@inproceedings{Proximal_denoiser,
  title={Proximal denoiser for convergent plug-and-play optimization with nonconvex regularization},
  author={Hurault, Samuel and Leclaire, Arthur and Papadakis, Nicolas},
  booktitle={International Conference on Machine Learning},
  pages={9483--9505},
  year={2022},
  organization={PMLR}
}

@inproceedings{PnP_methods,
  title={Plug-and-play methods provably converge with properly trained denoisers},
  author={Ryu, Ernest and Liu, Jialin and Wang, Sicheng and Chen, Xiaohan and Wang, Zhangyang and Yin, Wotao},
  booktitle={International Conference on Machine Learning},
  pages={5546--5557},
  year={2019},
  organization={PMLR}
}

@ARTICLE{MAP,
  author={Wipf, D.P. and Rao, B.D.},
  journal={IEEE Trans. Signal Process.}, 
  title={Sparse Bayesian learning for basis selection}, 
  year={2004},
  volume={52},
  number={8},
  pages={2153-2164},
  doi={10.1109/TSP.2004.831016}}

@book{pattern_recognition,
  title={Pattern recognition and machine learning},
  author={Bishop, Christopher M and Nasrabadi, Nasser M},
  volume={4},
  number={4},
  year={2006},
  publisher={Springer}
}

@ARTICLE{Env_aware_CE_using_FM,
  author={Liang, Guangming and Yang, Mingjie and Liu, Dongzhu and Henderson, Paul and Hanzo, Lajos},
  journal={IEEE Trans. Mobile Comput.}, 
  title={Environment-Aware Channel Inference via Cross-Modal Flow: From Multimodal Sensing to Wireless Channels}, 
  year={2026},
  volume={},
  number={},
  pages={1-17},
  doi={10.1109/TMC.2026.3717110}}

@INPROCEEDINGS{FM_for_CE_and_Symbol_detection,
  author={Guo, Yiyang and Qiao, Jing and Ye, Hao},
  booktitle={ICC 2026 - IEEE International Conference on Communications}, 
  title={Flow Matching for Joint Channel Estimation and Symbol Detection in Massive {MIMO}}, 
  year={2026},
  volume={},
  number={},
  pages={1-6},
  doi={10.1109/ICC59461.2026.11587708}}

@misc{emon2026flowsemflowmatchingadaptive,
      title={{FlowSem}: Flow Matching for Adaptive Wireless Image Transmission in Semantic Communication}, 
      author={Arnab Emon and Eslam Eldeeb},
      year={2026},
      eprint={2608.21651},
      archivePrefix={arXiv},
      primaryClass={eess.IV},
      url={https://arxiv.org/abs/2608.21651}, 
}

@misc{jia2025radioflowefficientradiomap,
      title={{RadioFlow}: Efficient Radio Map Construction Framework with Flow Matching}, 
      author={Haozhe Jia and Wenshuo Chen and Xiucheng Wang and Nan Cheng and Hongbo Zhang and Kuimou Yu and Songning Lai and Nanjian Jia and Bowen Tian and Hongru Xiao and Yutao Yue},
      year={2025},
      eprint={2510.09314},
      archivePrefix={arXiv},
      primaryClass={cs.CV},
      url={https://arxiv.org/abs/2510.09314}, 
}

@ARTICLE{11623723,
  author={Liang, Guangming and Yang, Mingjie and Liu, Dongzhu and Henderson, Paul and Hanzo, Lajos},
  journal={IEEE Trans. Mobile Comput.}, 
  title={Environment-Aware Channel Inference via Cross-Modal Flow: From Multimodal Sensing to Wireless Channels}, 
  year={2026},
  volume={},
  number={},
  pages={1-17},
  doi={10.1109/TMC.2026.3717110}}

@misc{mmWave_dataset,
  author       = {KIM, Minseok},
  title        = {{mm-Wave} Dual-Band Channel Sounding Dataset},
  month        = jul,
  year         = 2026,
  publisher    = {Zenodo},
  doi          = {10.5281/zenodo.21109349},
  url          = {https://doi.org/10.5281/zenodo.21109349},
}

\end{document}